\documentclass{article}

\PassOptionsToPackage{numbers,compress}{natbib}
\usepackage[preprint]{neurips_2026}

\usepackage[utf8]{inputenc}
\usepackage[T1]{fontenc}
\usepackage{hyperref}
\usepackage{url}
\usepackage{xurl}
\usepackage{booktabs}
\usepackage{amsfonts}
\usepackage{amsmath}
\usepackage{amssymb}
\usepackage{nicefrac}
\usepackage{microtype}
\usepackage{xcolor}
\usepackage{graphicx}
\usepackage{subcaption}
\usepackage{multirow}
\usepackage{multicol}
\usepackage{tabularx}
\usepackage{pifont}
\usepackage{float}
\usepackage{enumitem}
\usepackage{tcolorbox}
\usepackage{tikz}
\usetikzlibrary{arrows.meta,positioning,calc,shapes.geometric}
\usepackage{wasysym}

\definecolor{markgreen}{RGB}{0,170,0}
\definecolor{xmarkred}{RGB}{220,0,0}
\newcommand{\cmark}{\textcolor{markgreen}{\ding{51}}}
\newcommand{\xmark}{\textcolor{xmarkred}{\ding{55}}}
\definecolor{partialmark}{RGB}{139,69,19}
\newcommand{\pmark}{\textcolor{partialmark}{\LEFTcircle}}

\newcommand{\rs}[2]{#1{\scriptsize$\pm$#2}}
\newcommand{\rsb}[2]{\textbf{#1}{\scriptsize$\pm$#2}}

\providecommand{\dense}{\textsc{dense-opp}}
\providecommand{\sparse}{\textsc{sparse-opp}}

\title{FML-bench: A Controlled Study of AI Research Agent Strategies from the Perspective of Search Dynamics}

\author{%
  \textbf{Qiran Zou\textsuperscript{1,\,*}} \quad
  \textbf{Hou Hei Lam\textsuperscript{1,\,2,\,*}} \quad
  \textbf{Wenhao Zhao\textsuperscript{1}} \quad
  \textbf{Tingting Chen\textsuperscript{1}} \quad
  \textbf{Yiming Tang\textsuperscript{1}} \\
  \textbf{Samson Yu\textsuperscript{1}} \quad
  \textbf{Yingtao Zhu\textsuperscript{1}} \quad
  \textbf{Srinivas Anumasa\textsuperscript{1}} \quad
  \textbf{Zufeng Zhang\textsuperscript{2}} \quad
  \textbf{Tianyi Zhang\textsuperscript{3}} \\
  \textbf{Chang Liu\textsuperscript{2}} \quad
  \textbf{Zhengyao Jiang\textsuperscript{4}} \quad
  \textbf{Anirudh Goyal\textsuperscript{5}} \quad
  \textbf{Dianbo Liu\textsuperscript{1,\,\textdagger}} \\[6pt]
  \textsuperscript{1}National University of Singapore \quad
  \textsuperscript{2}Tsinghua University \quad
  \textsuperscript{3}University of Minnesota \\
  \textsuperscript{4}Weco \quad
  \textsuperscript{5}Meta
}

\begin{document}

\maketitle

\renewcommand{\thefootnote}{\fnsymbol{footnote}}
\footnotetext[1]{Equal contribution.}
\footnotetext[2]{Corresponding author. Contact: \texttt{qiranzou@u.nus.edu}, \texttt{dianbo@nus.edu.sg}.}
\renewcommand{\thefootnote}{\arabic{footnote}}
\setcounter{footnote}{0}


\begin{abstract}

AI research agents accelerate ML research by automating hypothesis generation, experimentation, and empirical refinement. Existing agent strategies range from greedy hill-climbing to tree search and evolutionary optimization, yet which strategy choices drive performance remains unclear.
Answering this question requires a benchmark that separates agent strategy (e.g., search topology) from execution infrastructure (e.g., code editor), so that performance differences are attributable to strategy rather than infrastructure, and that provides process-level metrics beyond final scores to analyze exploration behaviors.
Existing benchmarks offer limited support. We propose \textbf{FML-Bench}, a benchmark of 18 fundamental ML research tasks across 10 domains that separates agent strategy from execution infrastructure and defines 12 process-level behavioral metrics.
Evaluating six representative agents, we find that: (1) strategy complexity alone does not guarantee strong performance: a simple greedy hill-climber nearly matches the best-performing tree-search agent, both well above the remaining agents;
(2) our analysis suggests this pattern relates to improvement opportunity structure: greedy search tends to be more effective when opportunities are dense, while tree-search and evolutionary strategies tend to be more effective when opportunities are sparse; an adaptive agent built on this insight switches to broader exploration upon detecting improvement stagnation and outperforms the other six agents, lending initial support to this observation;
and (3) process-level analysis reveals that early convergence and directionally focused exploration are significantly associated with final performance, while solution diversity and compute cost are not.
Our benchmark is available at: \url{https://github.com/qrzou/FML-bench}.
\end{abstract}


\section{Introduction}
\label{sec:intro}
 
Automatic AI research agents show promise for accelerating ML research by automating the iterative loop of hypothesis generation, algorithm implementation, and experimentation. Agent strategies are evolving rapidly, now spanning a wide range of search topologies: from greedy hill-climbing~\cite{karpathy2026autoresearch} and parallel linear idea chains~\cite{lu2024ai}, through tree search~\cite{yamada2025ai,aide2025,toledo2025aira,chen2026mars}, to evolutionary optimization~\cite{openevolve}. New designs appear in quick succession, but evaluations lack controlled comparison, and a basic question remains: which aspects of agent strategy actually affect research outcomes?
 
Answering this question empirically requires a benchmark that satisfies two conditions. First, it should separate \emph{agent strategy} (how the agent generates ideas, selects what to try next, and learns from prior experiments) from \emph{execution infrastructure} (the code editor, experiment runner, and validation/test separation), so that performance differences are attributable to strategy rather than infrastructure. Second, it should provide process-level metrics beyond final scores, so that agents' exploration behaviors can be analyzed and related to outcomes.
Existing benchmarks offer limited support for these conditions: each agent typically ships with its own execution infrastructure, especially its own code editor, making it difficult to attribute performance differences to strategy alone; and existing evaluations largely report only final scores, compressing most of the research trajectory into a single scalar. Table~\ref{tab:benchmark_comparison} compares existing benchmarks along these axes.
 
We introduce FML-bench, comprising 18 fundamental ML research tasks across 10 domains (generalization, data efficiency, representation learning, continual learning, causality, robustness, privacy, fairness, unlearning, and federated learning), each built on a recognized baseline in a publicly available research codebase. FML-bench standardizes the execution infrastructure while preserving each agent's own strategy. In addition, FML-bench defines twelve process-level metrics that diagnose how each agent traverses the solution space.
 
\begin{figure}[t]
\centering
\includegraphics[width=\linewidth]{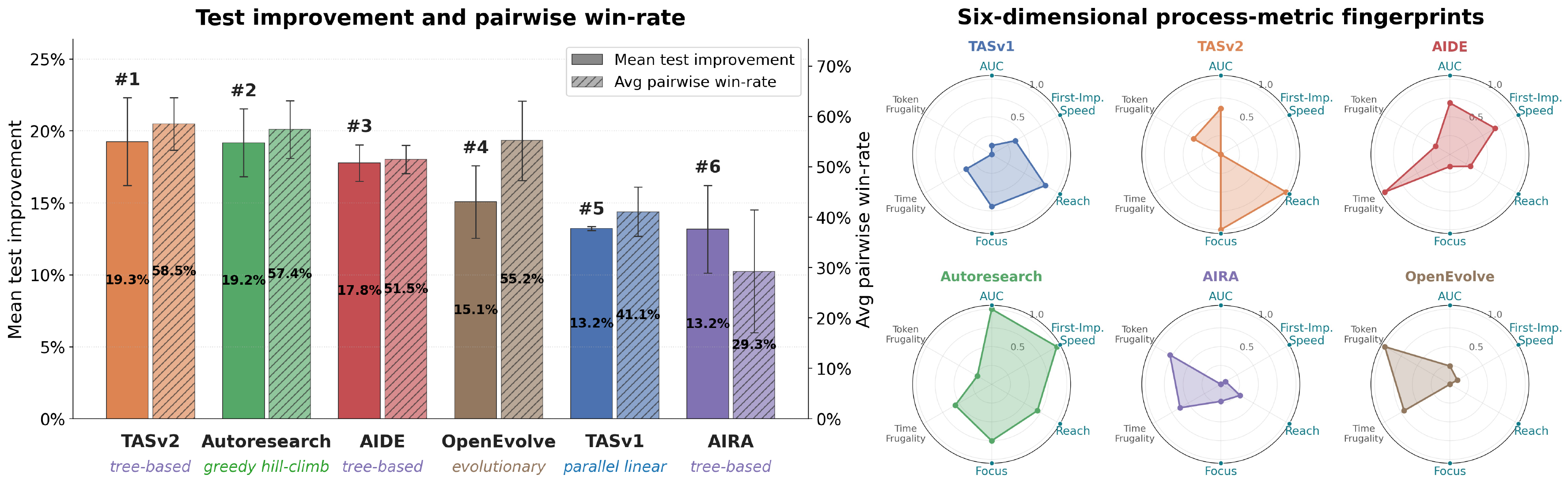}
\caption{\textbf{Comparison of the six AI research agents on FML-bench.} \emph{Left:} per-agent mean normalized test improvement (left axis) and average pairwise win-rate (right axis), agents ranked by mean improvement. \emph{Right:} per-agent fingerprint over six process-level axes capturing convergence efficiency, exploration geometry, and cost frugality (higher is better on every axis).}
\label{fig:six_agents_comparison}
\end{figure}
 
\textbf{Finding 1: Strategy complexity alone does not guarantee strong performance.}
Figure~\ref{fig:six_agents_comparison} previews the results of evaluating six AI research agents on FML-bench under the experimental protocol described in Section~\ref{sec:setup}. Autoresearch (a greedy hill-climber) and TAS~v2 (best-first tree search with LLM-journal memory) reach near-identical top-tier mean improvement and comparable pairwise win-rates, well above the remaining four agents. Despite the large gap in strategy complexity between the two, they achieve nearly the same aggregate performance, indicating that complexity alone does not determine outcomes.
 
\textbf{Finding 2: Greedy search tends to be more effective when improvement opportunities are dense; broader strategies tend to be more effective when opportunities are sparse.}
Investigating further, we find that Autoresearch excels on tasks where small code modifications frequently yield gains, but stalls on tasks where most modifications fail to improve the metric. Intuitively, greedy search exploits dense improvement landscapes efficiently, while broader strategies that maintain multiple search frontiers are better positioned to break through sparse ones. As an initial test, we build a simple agent, AdaptiveSearch, which starts with greedy search and switches to multi-branch exploration upon detecting stagnation. It outperforms both TAS~v2 and Autoresearch in mean improvement and win-rate, lending preliminary support to this observation.
 
\textbf{Finding 3: Early convergence and directionally focused exploration are associated with performance; compute scale and solution diversity are not.}
A correlation analysis of the twelve process-level metrics against final test improvement identifies two groups. Metrics capturing early improvement speed, exploration distance from the baseline, and directional concentration of the search trajectory are significantly associated with final performance. Commonly assumed predictors such as semantic-cluster count, total token cost, and wall-clock time show no significant association.
 
Our contributions are:
\begin{enumerate}[leftmargin=*,nosep]
    \item We introduce \textbf{FML-bench}, comprising 18 fundamental ML research tasks across 10 domains with execution-infrastructure separation and twelve process-level behavioral metrics.
 
    \item We show that the simplest greedy hill-climber (Autoresearch) matches a tree-search agent (TAS~v2) in aggregate performance, both well above the remaining four agents, indicating that strategy complexity alone does not guarantee performance.
 
    \item We find that greedy search tends to be more effective when improvement opportunities are dense, while broader strategies tend to be more effective when opportunities are sparse, and validate this observation with an adaptive switching agent.
 
    \item We identify early convergence and directionally focused exploration as significant predictors of final performance; solution diversity, token cost, and wall-clock time show no significant association.
\end{enumerate}

\section{Related Work}
\label{sec:related}

\begin{table}[t]
\caption{Comparison of different benchmarks. Exec.\ Infra.\ Ctrl.\ indicates whether the benchmark separates execution infrastructure (e.g., code editor) from agent strategy. \pmark\ denotes partial coverage: MLAgentBench and RE-Bench report token/cost and time only.}
\label{tab:benchmark_comparison}
\centering
\footnotesize
\begin{tabular}{llccc}
\toprule
\textbf{Benchmark} & \textbf{Task Type} & \textbf{\# Tasks} & \textbf{Process Metrics} & \textbf{Exec.\ Infra.\ Ctrl.} \\
\midrule
FML-bench (Ours) & Real-world ML research tasks & 18 & \cmark & \cmark \\
MLE-bench \cite{chan2024mle}            & Kaggle competitions                & 75   & \xmark & \xmark \\
MLAgentBench \cite{huang2023mlagentbench} & Handcrafted ML tasks \& Kaggle   & 13   & \pmark & \xmark \\
ML-Dev-Bench \cite{padigela2025ml}       & Handcrafted ML workflows          & 30   & \xmark & \xmark \\
SWE-bench \cite{jimenez2024swe}          & GitHub issue resolution            & 2294 & \xmark & \xmark \\
DSBench \cite{jing2024dsbench}           & ModelOff \& Kaggle competitions    & 540  & \xmark & \xmark \\
RE-Bench \cite{metr2024rebench}          & Handcrafted R\&D tasks             & 7    & \pmark & \xmark \\
\bottomrule
\end{tabular}
\end{table}

\subsection{AI research agents}
\label{sec:related_agents}

LLM-powered AI research agents span a growing space of agent strategies, which we organize here along the same search-topology axis used throughout this paper. At one end of the spectrum, Autoresearch~\cite{karpathy2026autoresearch} implements a greedy hill-climber that retains each code modification only when it strictly improves the validation metric. The AI Scientist v1~\cite{lu2024ai} generates a batch of research ideas up front and executes each independently with bounded retries, forming parallel linear chains without cross-idea feedback; its successor, The AI Scientist v2~\cite{yamada2025ai}, replaces this linear execution with best-first tree search, described below.

More structured search strategies span tree search and evolutionary optimization. The AI Scientist v2 performs best-first tree search over sequential research stages, guided by an LLM-maintained journal. AIDE~\cite{aide2025} also searches a solution tree but alternates between improving the best node and debugging failed leaves. AIRA~\cite{toledo2025aira} instantiates UCT-based MCTS with min--max fitness normalization, and MARS~\cite{chen2026mars} extends MCTS with budget-aware planning and a comparative reflective memory that distills cross-branch experimental lessons. At the evolutionary end, AlphaEvolve~\cite{novikov2025alphaevolve} maintains MAP-Elites-style island populations with LLM-guided mutation; OpenEvolve~\cite{openevolve}, an open-source reimplementation inspired by its publicly described design, serves as the population-based representative in our evaluation.
Beyond search topology, Agent Laboratory~\cite{schmidgall2025agent}, AI-Researcher~\cite{tang2025airesearcher}, and MLR-Copilot~\cite{li2024mlrcopilot} automate broader research workflows through multi-agent role specialization.

\subsection{Related benchmarks}
\label{sec:related_benchmarks}

Existing agent benchmarks span several task types but rarely focus on ML research settings. MLE-bench~\cite{chan2024mle} assembles 75 Kaggle competitions, and DSBench~\cite{jing2024dsbench} aggregates a similar portfolio from ModelOff and Kaggle. MLAgentBench~\cite{huang2023mlagentbench} provides 13 tasks drawn from standard ML datasets and Kaggle challenges, while ML-Dev-Bench~\cite{padigela2025ml} provides 30 handcrafted ML workflow tasks. SWE-bench~\cite{jimenez2024swe} focuses on GitHub issue resolution, and RE-Bench~\cite{metr2024rebench} evaluates frontier AI R\&D capability with score-over-time curves but does not standardize agent-side infrastructure.
In the broader AI-for-science space, ScienceAgentBench~\cite{chen2024scienceagentbench} evaluates data-science code generation, while SCIBENCH~\cite{wang2023scibench} and LAB-Bench~\cite{laurent2024labbench} test cross-disciplinary problem-solving and biology domain knowledge respectively; these target single-pass scientific tasks rather than the iterative ML research loop addressed here.
Across the ML-facing benchmarks listed above, each agent ships with its own execution infrastructure, especially its own code editor, making it difficult to attribute performance differences to strategy alone; and apart from MLAgentBench and RE-Bench, which report token cost and wall-clock time, existing evaluations report only final scores, compressing most of the search trajectory into a single scalar. Table~\ref{tab:benchmark_comparison} compares these benchmarks with FML-bench along both axes.


\section{FML-bench}
\label{sec:benchmark}

\begin{figure}[t]
\centering
\includegraphics[width=\linewidth]{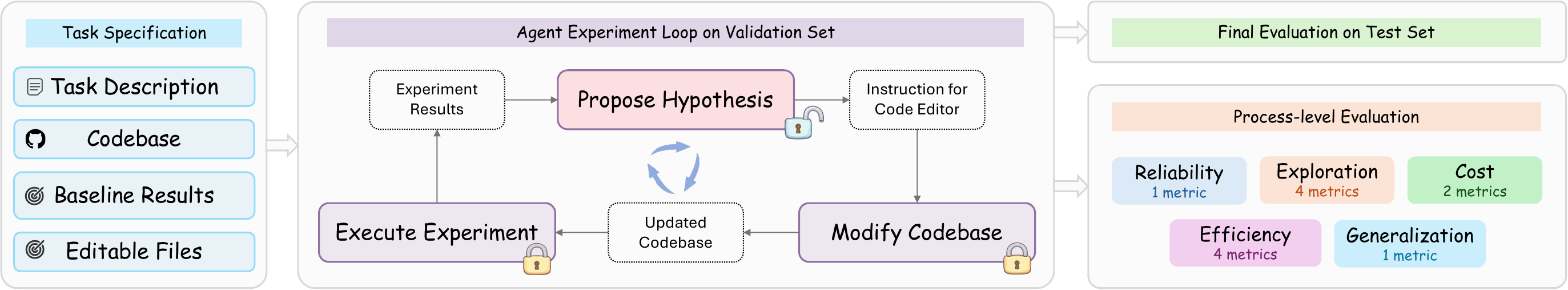}
\caption{\textbf{The FML-bench evaluation pipeline.} Left: the task specification fed to the agent. Center: the agent iterates a propose, modify, execute loop; only the decision of what to try next is governed by the agent's own strategy (unlocked icon), while codebase modification and experiment execution (locked icons) are shared framework infrastructure. Right: the framework evaluates the best-validated codebase on a held-out test set and records process-level metrics.}
\label{fig:fml_pipeline}
\end{figure}

\subsection{Task design}
\label{sec:task_design}

FML-bench comprises 18 research tasks spanning 10 machine-learning domains that range from system trustworthiness (generalization, robustness, privacy, fairness, unlearning)~\cite{li2023trustworthy, mehrabi2021survey} to learning capability (data efficiency, representation learning, continual learning, causality, federated learning)~\cite{bengio2013representation, pearl2019seven, chen2018lifelong}. Each task is built on an established baseline in a publicly available research codebase, and the agent is asked to improve the baseline. To keep the benchmark accessible, we require that a single validation run of each completes within 40 minutes on one GPU, so that a full 100-step agent run finishes within a few days. Per-task details, including datasets, baselines, and metrics, are given in Appendix~\ref{app:tasks}.

\subsection{Evaluation pipeline}
\label{sec:controlled_framework}

FML-bench isolates agent strategy as the sole independent variable by separating it from execution infrastructure. Each agent retains its own agent strategy (search topology, prompt templates, memory mechanism, and action vocabulary), collectively governing how the agent generates ideas, selects what to try next, and learns from prior experiments. The surrounding execution infrastructure, including code editing, experiment execution, metric display, and validation/test separation, is shared across all agents and provided by the benchmark framework, so that performance differences are attributable to agent strategy rather than to infrastructure differences.

\paragraph{Iterative experiment loop.}
Figure~\ref{fig:fml_pipeline} illustrates the evaluation pipeline. A task specification (left panel) supplies the agent with a fixed task description, an initial codebase, baseline validation results, and a pre-declared list of editable files. The agent then enters an iterative loop (center panel): it proposes an idea based on prior experiment results, the resulting edit instruction is passed to the framework's shared code editor for codebase modification, and the modified codebase is run through the framework's shared execution environment to produce experiment results that feed back into the next idea. Within this loop, only the decision of what to try next is governed by the agent's own strategy (unlocked icon in Figure~\ref{fig:fml_pipeline}); codebase modification and experiment execution (locked icons) are framework-controlled infrastructure identical for every agent. After the step budget is exhausted, the framework selects the best-validated codebase and runs a single held-out test evaluation (right panel), recording both the test metric and process-level metrics for behavioral analysis.

\paragraph{Shared infrastructure components.}
The shared infrastructure comprises four components.
\textbf{(1)}~The shared code editor is a thin wrapper that issues a single LLM API call per edit and logs the full prompt and patch; each agent passes its own instruction string, but no agent embeds additional tool-use scaffolding, compiler feedback loops, or retrieval augmentation inside the editor, so there is no agent-specific intelligence in the editing layer.
\textbf{(2)}~The unified step definition counts one step as one execution of the task's validation command; LLM calls internal to idea generation, journal maintenance, or code editing do not advance the step counter, so agents that deliberate more per step are not penalized by the step budget.
\textbf{(3)}~Framework-enforced val/test separation ensures that the agent observes only validation metrics throughout the search; after the budget is exhausted, the framework evaluates the best-validated codebase on a fully held-out test set whose inputs, outputs, and feedback are never exposed to the agent or any of its internal modules.
\textbf{(4)}~The unified metric display renders every metric value through a single formatting routine that emits the task's primary metric, its direction of improvement, and any constraint-violation flag in one identical format, so that metric parsing cannot leak hidden advantage.

\subsection{Final performance metrics}
\label{sec:final_metrics}

We define two complementary metrics to evaluate agent performance at the end of each run.

\paragraph{Normalized test improvement.}
Let $p_{\text{agent}}$ denote the raw test metric of the best-validated codebase, evaluated on the held-out test set at the end of the run (Section~\ref{sec:controlled_framework}). We report performance as a normalized test improvement
$\hat{p} = \max\!\bigl(0,\; (p_{\text{agent}} - p_{\text{baseline}}) / |p_{\text{best}} - p_{\text{worst}}|\bigr)$, where $p_{\text{baseline}}$ is the untouched baseline metric and $p_{\text{best}}$, $p_{\text{worst}}$ are the theoretical best and worst values of the task's primary metric (e.g.\ $1.0$ and $0.0$ for accuracy). Improvements below the baseline are clamped to~$0$, and the denominator places tasks on a common scale. The complete specification, including the convention for lower-is-better metrics and tasks with unbounded worst values, is given in Appendix~\ref{app:metrics}.

\paragraph{Pairwise win-rate.}
For each task, agent $A$ strictly wins against agent $B$ if its raw test metric is strictly better than $B$'s in the task's native direction; ties are not counted as wins. Each agent's overall win-rate is the average of its strict-win fraction against every other agent across all tasks; the formal definition is given in Appendix~\ref{app:metrics}. This rank-based metric is invariant to monotone rescaling of the test metric and complements the normalized improvement.

\subsection{Process-level metrics}
\label{sec:metrics}

To diagnose how each agent traverses the solution space, FML-bench defines twelve scalar process-level metrics organized along five analysis dimensions, each capturing a distinct hypothesis about what predicts agent performance. All twelve metrics are summarized below by dimension; complete formal specifications, including edge-case conventions, are given in Appendix~\ref{app:metrics}.

\paragraph{Exploration.}
These metrics measure how widely and how directionally the agent searches in code space. Let $\mathbf{e}_t$ denote the GraphCodeBERT~\cite{guo2020graphcodebert} embedding of the full codebase at the $t$-th valid step, $\bar{\mathbf{e}} = \frac{1}{n}\sum_{t} \mathbf{e}_t$ the trajectory centroid, $\mathbf{e}_\text{base}$ the baseline embedding, $n$ the number of valid steps, and $\{\lambda_i\}$ the eigenvalues of the sample covariance of the centered embeddings. GraphCodeBERT operates over the program's data-flow graph, so its embeddings reflect structural code change rather than surface token substitution. The four exploration metrics are:
\begin{equation}
\label{eq:exploration}
\begin{aligned}
D_{\text{spread}} &= \tfrac{1}{n}\textstyle\sum_{t=1}^{n}\left\| \mathbf{e}_t - \bar{\mathbf{e}} \right\|_2, &\qquad
D_{\text{reach}} &= \max_{t \in \{1,\ldots,n\}} \left\| \mathbf{e}_t - \mathbf{e}_\text{base} \right\|_2, \\[4pt]
U &= \tfrac{\#\text{clusters}(\{\mathbf{e}_t\})}{n}, &\qquad
D_{\text{eff}} &= \tfrac{\left(\sum_i \lambda_i\right)^2}{\sum_i \lambda_i^2}.
\end{aligned}
\end{equation}
\textbf{(a)}~Exploration Spread ($D_{\text{spread}}$) measures the mean within-trajectory dispersion around the centroid. \textbf{(b)}~Exploration Reach ($D_{\text{reach}}$) records the farthest point the trajectory ever reaches from the starting baseline. \textbf{(c)}~Exploration Uniqueness ($U$) captures how many distinct solution families the agent touches, via cosine agglomerative clustering on the embedding sequence. \textbf{(d)}~Effective dim ($D_{\text{eff}}$), the participation ratio, is low when exploration concentrates along a few principal directions and high when it spreads isotropically.

\paragraph{Generalization.}
The val-test gap is the absolute difference between the normalized validation and test improvements (each computed analogously to $\hat{p}$ in Section~\ref{sec:final_metrics}) on the best-validated codebase: $|\hat{p}_\text{val} - \hat{p}_\text{test}|$. 

\paragraph{Reliability.}
The valid step ratio $R_\text{valid} = n / T$ is the fraction of the $T = 100$ optimization steps that produced a valid result, i.e., neither an execution error, a timeout, nor a task-specific constraint violation.

\paragraph{Efficiency.}
These metrics capture when and how consistently gains accumulate over the step budget. \textbf{(a)}~AUC-over-steps is the time-averaged best-so-far normalized validation improvement, i.e.\ the area under the monotone envelope of the validation trajectory over $T$ steps; agents that improve earlier and hold their lead score higher. \textbf{(b)}~First-improvement step is the first step at which validation surpasses the baseline (undefined if this never occurs). \textbf{(c)}~Best-improvement step records when the run's peak validation was first achieved. \textbf{(d)}~Late-gain fraction measures the share of final improvement earned in the second half of the budget.

\paragraph{Cost.}
\textbf{(a)}~Token cost is the total number of prompt and completion tokens across all LLM calls during the run, including code-editing, idea-generation, and journal-maintenance calls. \textbf{(b)}~Wall-clock time is the total run duration in hours, from the first agent action to the final test evaluation.

\section{How Do Agent Strategies Compare?}
\label{sec:results_q1}

\subsection{Experimental Setup}
\label{sec:setup}

\paragraph{Agents.} We evaluate six AI research agents that span the contemporary strategy space: The AI Scientist v1~\cite{lu2024ai}, The AI Scientist v2~\cite{yamada2025ai}, AIDE~\cite{aide2025}, AIRA~\cite{toledo2025aira}, Autoresearch~\cite{karpathy2026autoresearch}, and OpenEvolve~\cite{openevolve}. The long-form description of each agent's search algorithm, with implementation parameters, stage budgets, and sampling rules, is in Appendix~\ref{app:agents}.

\paragraph{Protocol.} Every agent is run on every task for 3 independent rounds 
under a budget of $T = 100$ optimization steps per round (step definition in Section~\ref{sec:controlled_framework}), yielding $n = 324$
(round, agent, task) runs in total (6 agents $\times$ 18 tasks $\times$ 3 rounds). All 324 main-experiment runs share a single underlying LLM, GPT-5.4, so that any performance gap reflects agent strategy rather than LLM capability. All training, validation, and testing jobs execute on NVIDIA A100-80GB GPUs inside task-specific conda environments. 
Agent performance on a task is reported as the normalized test improvement $\hat{p}$ defined in Section~\ref{sec:final_metrics}.
The end-of-run test evaluation follows the pipeline of Section~\ref{sec:controlled_framework}.

\subsection{Per-agent performance and behavioral profiles}
\label{sec:crossover}

Across the 18 research tasks, TAS~v2 leads with the highest mean normalized test improvement ($0.193$) and pairwise win-rate ($58.5\%$); Autoresearch nearly matches it ($0.192$, $57.4\%$); AIDE and OpenEvolve sit in the middle tier; TAS~v1 and AIRA rank at the bottom (full per-task results in Table~\ref{tab:normalized_performance}).
Notably, the top position is held by the most complex strategy in the suite (a four-stage best-first tree search with LLM-journal memory), yet it is nearly matched by the simplest (a greedy hill-climber), while the most theoretically principled UCT-MCTS agent ranks last, indicating that strategy complexity does not monotonically predict performance.

The near-tie between the two leaders, however, masks markedly different search behaviors.
Table~\ref{tab:process_metrics} reports per-agent means for the twelve process-level metrics and reveals the following contrasts; the systematic relationship between these behavioral properties and final performance is analyzed in Section~\ref{sec:results_q2}.

\paragraph{Exploration behavior.}
TAS~v2 and Autoresearch both exhibit a ``far-and-focused'' exploration pattern (Reach ranked first and third; Effective dim ranked lowest and second-lowest), whereas OpenEvolve shows the opposite profile: highest Uniqueness yet lowest Reach, maintaining population diversity but searching close to the baseline throughout.
TAS~v1 has the second-highest Reach but explores in more scattered directions (higher Effective dim than the two leaders).

\paragraph{Efficiency behavior.}
Gain accumulation splits into two regimes.
Autoresearch and AIDE follow a ``fast-frontloaded'' pattern: earliest first-improvement steps and highest AUC-over-steps.
TAS~v2 and AIRA follow a ``slow-start'' pattern: latest first-improvement steps and highest late-gain fractions.
The dividing line is whether the second-half gains suffice: TAS~v2's late push closes the gap to the top, whereas AIRA's does not (lowest AUC overall).

\paragraph{Reliability and cost.}
AIDE achieves the highest valid-step ratio ($0.889$), TAS~v2 the lowest ($0.734$).
TAS~v1 consumes the most tokens ($2.29$M), OpenEvolve the fewest ($0.93$M); wall-clock time varies little across agents.

Section~\ref{sec:results_q2} systematically analyzes which behavioral properties predict final performance. Section~\ref{sec:autoresearch_deep_dive} investigates why the simplest greedy strategy can match the most complex tree-search strategy in aggregate, and validates the resulting insight through an adaptive agent.

\begin{table}[t]
\centering
\footnotesize
\caption{Normalized test improvement for each (agent, task) cell, aggregated as mean across 3 rounds; \rs{0.000}{.000} means the agent never surpassed the baseline in any round.}
\label{tab:normalized_performance}
\begin{tabular}{lcccccc}
\toprule
Task & TAS v1 & TAS v2 & AIDE & AIRA & AutoR & OEvolve \\
\midrule
DomainBed-CM & \rs{0.071}{.041} & \rsb{0.245}{.182} & \rs{0.231}{.198} & \rs{0.165}{.211} & \rs{0.067}{.053} & \rs{0.242}{.200} \\
DomainBed-OH & \rs{0.006}{.004} & \rsb{0.014}{.003} & \rs{0.011}{.005} & \rs{0.012}{.002} & \rs{0.005}{.004} & \rs{0.013}{.005} \\
EasyFSL      & \rs{0.013}{.020} & \rs{0.034}{.017} & \rs{0.030}{.046} & \rsb{0.063}{.025} & \rs{0.032}{.028} & \rs{0.046}{.028} \\
USB          & \rs{0.040}{.011} & \rs{0.039}{.008} & \rs{0.029}{.008} & \rs{0.019}{.005} & \rs{0.048}{.029} & \rsb{0.067}{.012} \\
Lightly      & \rs{0.020}{.025} & \rs{0.019}{.016} & \rs{0.044}{.023} & \rs{0.000}{.000} & \rsb{0.069}{.029} & \rs{0.043}{.012} \\
Solo-learn   & \rs{0.076}{.018} & \rsb{0.113}{.004} & \rs{0.069}{.025} & \rs{0.066}{.034} & \rs{0.003}{.003} & \rs{0.093}{.005} \\
Cont.-Learn. & \rs{0.048}{.011} & \rs{0.349}{.339} & \rs{0.113}{.163} & \rs{0.072}{.124} & \rsb{0.375}{.230} & \rs{0.164}{.089} \\
PyCIL        & \rs{0.030}{.025} & \rsb{0.064}{.004} & \rs{0.056}{.005} & \rs{0.020}{.007} & \rs{0.026}{.007} & \rs{0.037}{.021} \\
CausalML     & \rs{0.020}{.006} & \rsb{0.040}{.068} & \rs{0.023}{.024} & \rs{0.033}{.057} & \rs{0.007}{.012} & \rs{0.014}{.024} \\
gCastle      & \rs{0.014}{.014} & \rs{0.066}{.114} & \rs{0.127}{.207} & \rs{0.145}{.045} & \rs{0.127}{.143} & \rsb{0.169}{.146} \\
ART          & \rs{0.429}{.026} & \rs{0.395}{.026} & \rs{0.447}{.035} & \rs{0.303}{.127} & \rs{0.448}{.032} & \rsb{0.459}{.011} \\
OpenOOD      & \rs{0.023}{.033} & \rs{0.039}{.017} & \rs{0.012}{.020} & \rs{0.021}{.002} & \rsb{0.057}{.013} & \rs{0.006}{.011} \\
PrivacyMeter & \rs{0.008}{.014} & \rs{0.482}{.068} & \rs{0.530}{.018} & \rs{0.011}{.020} & \rsb{0.575}{.025} & \rs{0.032}{.056} \\
Opacus       & \rs{0.000}{.000} & \rs{0.000}{.000} & \rs{0.006}{.010} & \rs{0.000}{.000} & \rsb{0.054}{.017} & \rs{0.003}{.003} \\
AIF360       & \rs{0.208}{.028} & \rs{0.218}{.031} & \rsb{0.231}{.005} & \rs{0.136}{.120} & \rs{0.218}{.016} & \rs{0.156}{.135} \\
Fairlearn    & \rs{0.170}{.005} & \rsb{0.173}{.000} & \rs{0.162}{.003} & \rs{0.152}{.002} & \rs{0.170}{.001} & \rsb{0.173}{.000} \\
Unlearning   & \rsb{0.968}{.004} & \rs{0.944}{.016} & \rs{0.831}{.142} & \rs{0.921}{.045} & \rs{0.896}{.094} & \rs{0.794}{.039} \\
PFLlib       & \rs{0.236}{.018} & \rs{0.236}{.013} & \rs{0.250}{.007} & \rs{0.234}{.042} & \rsb{0.277}{.053} & \rs{0.201}{.005} \\
\midrule
MEAN & \rs{0.132}{.236} & \rsb{0.193}{.237} & \rs{0.178}{.222} & \rs{0.132}{.215} & \rs{0.192}{.243} & \rs{0.151}{.197} \\
\bottomrule
\end{tabular}
\end{table}

\subsection{Greedy search excels on opportunity-dense tasks; broader exploration suits opportunity-sparse tasks}
\label{sec:autoresearch_deep_dive}

\begin{table}[t]
\centering
\footnotesize
\caption{Seven-agent comparison including AdaptiveSearch. Mean normalized test improvement and pairwise win-rate across 18 tasks, agents ranked by mean improvement.}
\label{tab:adaptive_comparison}
\setlength{\tabcolsep}{4pt}
\begin{tabular}{lccccccc}
\toprule
& Adaptive (ours) & TAS v2 & AutoR & AIDE & OEvolve & TAS v1 & AIRA \\
\midrule
Mean impr. & \textbf{0.208} & 0.193 & 0.192 & 0.178 & 0.151 & 0.132 & 0.132 \\
Win-rate   & \textbf{58.6\%} & 56.2\% & 56.2\% & 49.4\% & 52.8\% & 40.4\% & 28.4\% \\
\bottomrule
\end{tabular}
\end{table}

Section~\ref{sec:crossover} reveals a seemingly paradoxical result: Autoresearch, a greedy hill-climber, nearly matches TAS~v2, the most complex strategy in the suite, on overall ranking. If broader exploration is generally considered advantageous in search theory, why does the simplest greedy strategy achieve comparable aggregate performance?

We hypothesize that the effectiveness of a search strategy depends on the task's \emph{improvement opportunity structure}. When improvement opportunities are dense, meaning that small code modifications have a high probability of raising the validation metric, greedy hill-climbing can efficiently exploit these opportunities step by step without spending budget on uncertain new directions. When improvement opportunities are sparse, meaning that most modifications fail to produce gains and structurally different approaches are needed to break through plateaus, greedy search tends to stall because it immediately discards candidates that do not strictly improve the validation metric; tree-search and evolutionary strategies, by maintaining multiple search frontiers, are more likely to discover directions that break through such plateaus. If this hypothesis holds, an agent that can judge online when greedy search is no longer effective and adaptively switch to broader exploration should outperform both.

To test this hypothesis, we design AdaptiveSearch. Its core observation is that if greedy search has not produced any meaningful improvement in the validation metric over a sustained window of recent steps, this signals that greedy has stalled and the strategy should switch. This signal is fully computable online and does not rely on any cross-task or cross-agent external metric. Based on this, AdaptiveSearch adaptively switches search strategies within a single $T{=}100$ step budget:
\begin{itemize}[leftmargin=*,nosep]
    \item \textbf{Phase~1} (greedy) is identical to Autoresearch: a single incumbent is maintained, and a modification is accepted only when validation strictly improves; otherwise the change is rolled back.
    \item \textbf{Transition criterion}: when the normalized improvement has not increased over the most recent $W$ consecutive steps, the agent irreversibly enters Phase~2.
    \item \textbf{Phase~2} (multi-branch exploration) forks multiple independent branches from the top-$N$ candidates accumulated during Phase~1 and advances them in round-robin fashion (details in Appendix~\ref{app:adaptive}).
\end{itemize}
The key design idea is to make no upfront assumption about which regime a task belongs to, but instead use greedy search as a probe: as long as greedy continues to improve, the agent exploits its efficiency; once greedy stalls, the agent switches to broader exploration.

As shown in Table~\ref{tab:adaptive_comparison}, AdaptiveSearch achieves a mean normalized test improvement of $0.208$ on the same 18 tasks under the same 100-step budget, above both TAS~v2 ($0.193$) and Autoresearch ($0.192$); its pairwise win-rate of $58.6\%$ likewise exceeds TAS~v2 ($56.2\%$) and Autoresearch ($56.2\%$).
A post-hoc partition of the 18 tasks by opportunity density (Appendix~\ref{app:autoresearch_polarization}) further supports this conclusion: Autoresearch ranks first on high-density tasks but drops to sixth on low-density tasks, while AdaptiveSearch ranks in the top two on both partitions.

This result validates the hypothesis on two levels. First, AdaptiveSearch outperforms both the purely greedy Autoresearch and the broad-exploration TAS~v2, confirming that the two strategies have complementary advantages on different types of tasks. Second, a simple online signal, whether the validation metric has stalled over recent consecutive steps, suffices to determine when to switch from greedy to broader exploration, showing that adaptive search is feasible without prior knowledge of a task's improvement opportunity structure. This provides a practical guideline for agent strategy design: search strategies should not be fixed, but should adapt based on the online-observed improvement dynamics.

\begin{table}[t]
\centering
\footnotesize
\setlength{\tabcolsep}{4pt}
\caption{Process metrics summary: per-agent means and pooled Spearman correlation between each metric and normalized test improvement. Significant associations ($p < 0.05$) are shown in bold.}
\label{tab:process_metrics}
\begin{tabular}{llrrrrrrrr}
\toprule
Dimension & Metric & TAS v1 & TAS v2 & AIDE & AIRA & AutoR & OEvolve & $\rho$ & $p$ \\
\midrule
Exploration & Exploration Spread & 23.65 & 28.44 & 16.23 & 14.60 & 23.35 & 10.89 & +0.094 & .091 \\
Exploration & Exploration Uniqueness & 0.0628 & 0.0474 & 0.0357 & 0.0406 & 0.0341 & 0.0796 & +0.012 & .823 \\
Exploration & Exploration Reach & 122.6 & 139.6 & 72.97 & 71.19 & 110.6 & 42.39 & \textbf{+0.115} & \textbf{.039} \\
Exploration & Effective dim & 1.784 & 1.450 & 2.974 & 2.745 & 1.709 & 3.718 & \textbf{$-$0.140} & \textbf{.011} \\
Generalization & Val-test $|$gap$|$ & 0.0362 & 0.0247 & 0.0274 & 0.0401 & 0.0334 & 0.0448 & +0.097 & .080 \\
Reliability & Valid step ratio & 0.7744 & 0.7339 & 0.8887 & 0.7893 & 0.8241 & 0.8281 & +0.086 & .122 \\
Efficiency & AUC-over-steps & 0.1356 & 0.1579 & 0.1612 & 0.1303 & 0.1755 & 0.1412 & \textbf{+0.784} & $\mathbf{<}$\textbf{.001} \\
Efficiency & First-improvement step & 12.78 & 16.46 & 10.62 & 15.56 & 9.185 & 15.11 & \textbf{$-$0.291} & $\mathbf{<}$\textbf{.001} \\
Efficiency & Late-gain fraction & 0.1067 & 0.3016 & 0.1185 & 0.3205 & 0.1882 & 0.2246 & $-$0.042 & .462 \\
Efficiency & Best-improvement step & 48.70 & 74.17 & 79.09 & 71.78 & 91.24 & 70.94 & +0.104 & .060 \\
Cost & Token cost (M) & 2.288 & 1.426 & 1.737 & 1.075 & 1.736 & 0.9343 & $-$0.005 & .935 \\
Cost & Wall-clock time (h) & 30.27 & 32.49 & 27.35 & 29.08 & 29.41 & 28.70 & $-$0.094 & .090 \\
\bottomrule
\end{tabular}
\end{table}


\section{What Predicts Agent Success?}\label{sec:results_q2}

Section~\ref{sec:crossover} characterized each agent's behavioral profile and revealed that agents with similar aggregate rankings can arrive there through markedly different search behaviors. This section asks which behavioral properties actually predict final performance: we first qualitatively align process-metric rankings with agent performance rankings (Section~\ref{sec:alignment}), then quantitatively verify these observations with pooled Spearman rank correlations (Section~\ref{sec:correlation_analysis}).

\subsection{Behavioral properties versus performance ranking}\label{sec:alignment}

\paragraph{Efficiency metrics align closely with performance ranking.}
The per-agent ordering of AUC-over-steps matches the final-performance ranking most closely: the top three on AUC (Autoresearch, AIDE, TAS~v2) are exactly the top three on final test improvement, and the bottom two on AUC (TAS~v1, AIRA) are likewise the bottom two on final performance. First-improvement step shows a similar trend, with one exception: TAS~v2 starts slowest yet still ranks first overall, its efficiency disadvantage compensated by exploration strengths.

\paragraph{Two paths to the top.}
TAS~v2 compensates for its slow start with the highest Exploration Reach and the lowest Effective dim, searching far from the baseline along few principal directions. Autoresearch compensates for its more modest exploration distance with the highest AUC and the fastest first improvement. Final performance is not determined by any single behavioral dimension but by different combinations of exploration geometry and efficiency.

\paragraph{Broad exploration does not guarantee high performance.}
TAS~v1 provides a counterexample: it ranks second on both Reach and Spread, yet finishes fifth on final performance. Compared with TAS~v2, its exploration is more directionally scattered (higher Effective dim) and its gains fail to accumulate (lower AUC), indicating that the directional quality of exploration matters more than its breadth.

\paragraph{Cost is decoupled from performance.}
The most expensive agent (TAS~v1) ranks fifth; the cheapest (OpenEvolve) ranks fourth. The agent with the highest valid-step ratio (AIDE) ranks third, while the one with the lowest (TAS~v2) ranks first.

\subsection{Correlation analysis}\label{sec:correlation_analysis}

\paragraph{Significant predictors.}
Pooled Spearman rank correlations over the 324 (round, agent, task) runs (last two columns of Table~\ref{tab:process_metrics}) confirm the qualitative observations of Section~\ref{sec:alignment}; four metrics reach significance ($p < 0.05$).
\textbf{(1)}~AUC-over-steps is the strongest single predictor ($\rho = +0.784$, $p = 1.6 \times 10^{-68}$), confirming the efficiency--ranking alignment; part of this strength likely reflects definitional overlap between cumulative and final improvement, but the magnitude far exceeds what the overlap alone would produce.
\textbf{(2)}~First-improvement step is negatively correlated ($\rho = -0.291$, $p = 2.3 \times 10^{-7}$): agents that surpass the baseline earlier tend to finish higher, consistent with the fast-start pattern of Autoresearch and AIDE.
\textbf{(3)}~Exploration Reach is positively correlated ($\rho = +0.115$, $p = 0.039$), providing statistical support for TAS~v2's reach-driven path to the top.
\textbf{(4)}~Effective dim is negatively correlated ($\rho = -0.140$, $p = 0.011$), validating the observation that TAS~v1's broad but directionally scattered exploration does not translate into performance.

\paragraph{Non-significant predictors.}
The remaining eight metrics do not reach significance on the pooled sample; we highlight three that are commonly assumed to matter.
\textbf{(1)}~Exploration Uniqueness shows essentially no correlation ($\rho = +0.012$, $p = 0.82$). Combined with the significant positive correlation of Exploration Reach, this indicates that exploration quality is better measured by how far the trajectory travels from the baseline than by how many distinct neighborhoods it visits, providing empirical nuance to the novelty-search~\cite{lehman2011abandoning} and quality-diversity~\cite{pugh2016quality,mouret2015illuminating} literature.
\textbf{(2)}~Neither token cost ($\rho = -0.005$, $p = 0.94$) nor wall-clock time ($\rho = -0.094$, $p = 0.090$) correlates with test improvement; under a fixed 100-step budget, how an agent allocates its compute matters more than how much it spends.

\section{Conclusion}
\label{sec:conclusion}

We introduced FML-bench, comprising 18 fundamental ML research tasks across 10 domains, and used it to conduct a controlled comparison of six representative AI research agents by separating execution infrastructure from agent strategy and defining twelve process-level behavioral metrics. We find that strategy complexity alone does not guarantee performance; greedy search tends to be more effective when improvement opportunities are dense while broader strategies tend to be more effective when opportunities are sparse; and early convergence speed and directionally focused exploration are significantly associated with final performance, while solution diversity and compute cost are not. The main limitations of this work are that, to enable controlled comparison, we uniformly removed each agent's native code editor and auxiliary subsystems (Appendix~\ref{app:porting}), which may underestimate each agent's performance under its full original configuration; six agents cover the major search topologies but are not exhaustive; and the fixed step budget may affect relative rankings. We hope that FML-bench can serve as a controlled experimental foundation for evidence-driven agent strategy design, and see online strategy switching and broader agent coverage as natural next steps.

\bibliographystyle{plainnat}
\bibliography{references}

\newpage
\appendix
\section{Task descriptions}\label{app:tasks}

This appendix gives a short description of each of the 18 research tasks in FML-bench, one paragraph per task. For every task we identify the dataset, the baseline algorithm, the agent's optimization target, and the source repository. Table~\ref{tab:task_cards} collects the same information in a single reference card, together with the baseline raw metric, the per-task opportunity density $\phi_\text{opp}$, and the empirical \dense{} / \sparse{} partition assignment from Section~\ref{app:autoresearch_polarization}.

\paragraph{Generalization (DomainBed, ColoredMNIST).}
ColoredMNIST~\cite{arjovsky2019invariant} is a controlled distribution-shift benchmark in which digit color is spuriously correlated with the label and is inverted on a held-out test domain. The baseline is Empirical Risk Minimization~\cite{Vapnik1998} trained on the two source domains, and the agent is asked to raise accuracy on the held-out test domain. Repository: DomainBed~\cite{gulrajani2020search}.

\paragraph{Generalization (DomainBed, OfficeHome).}
OfficeHome~\cite{venkateswara2017deep} presents a realistic four-domain image-classification problem (Art, Clipart, Product, Real) built on a ResNet-50 backbone; the baseline is ERM with pretrained ResNet-50 features~\cite{gulrajani2020search}. The agent must raise the leave-one-domain-out average accuracy across all four domains. Repository: DomainBed~\cite{gulrajani2020search}.

\paragraph{Data Efficiency (EasyFSL, Mini-ImageNet).}
EasyFSL~\cite{SicaraEasyFSL} provides a lightweight few-shot classification harness on Mini-ImageNet~\cite{vinyals2016matching}, whose fine-grained inter-class similarity makes it a standard few-shot benchmark. The baseline is Prototypical Networks~\cite{snell2017prototypical} operating over a frozen backbone, and the agent is asked to raise few-shot classification accuracy.

\paragraph{Data Efficiency (USB, FixMatch).}
The USB framework~\cite{wang2022usb} supplies a unified semi-supervised learning harness; the task is CIFAR-100 with only 200 labeled examples, and the baseline is FixMatch~\cite{sohn2020fixmatch}. The agent must raise test accuracy by better exploiting the large pool of unlabeled data.

\paragraph{Representation Learning (Lightly, MoCo).}
Lightly~\cite{lightly} implements modern self-supervised learning on CIFAR-10~\cite{krizhevsky2009learning}; the baseline is MoCo~\cite{he2020momentum}, evaluated by training a linear classifier on frozen features. The agent must raise linear-probe accuracy by improving the pretraining stage.

\paragraph{Representation Learning (Solo-learn, Barlow Twins).}
Solo-learn~\cite{da_costa2022sololearn} provides a library of self-supervised methods on CIFAR-100; the baseline is Barlow Twins~\cite{zbontar2021barlowtwins}, evaluated by standard linear evaluation on frozen features. The agent must raise linear-eval accuracy.

\paragraph{Continual Learning (Continual-Learning, SI on splitMNIST).}
This task mitigates catastrophic forgetting on a class-incremental split of MNIST~\cite{deng2012mnist}; the baseline is Synaptic Intelligence~\cite{zenke2017continual}, and no memory replay buffer is allowed. The agent must raise average accuracy across all tasks at the end of training. Repository: the continual-learning codebase of~\citet{vandeven2022three}.

\paragraph{Continual Learning (PyCIL, iCaRL on CIFAR-100).}
PyCIL~\cite{zhou2023pycil} is a standard class-incremental learning toolbox on CIFAR-100; the baseline is iCaRL~\cite{rebuffi2017icarl}, and the target metric is average incremental accuracy across the sequential-task curriculum. The agent must raise this average.

\paragraph{Causality (CausalML, Dragonnet on IHDP).}
CausalML~\cite{chen2020causalml} provides treatment-effect estimators on IHDP~\cite{hill2011bayesian}; the baseline is Dragonnet~\cite{shi2019adapting}, and the agent must reduce the mean absolute error of individual treatment-effect predictions. Because MAE has no natural upper bound, this task uses the $p_\text{worst} = p_\text{baseline}$ normalization convention of Appendix~\ref{app:metrics}.

\paragraph{Causality (gCastle, NOTEARS).}
gCastle~\cite{zhang2023gcastle} implements continuous-optimization causal discovery; the baseline is NOTEARS~\cite{zheng2018dags} on nonlinear 50-node synthetic DAGs. The agent must reduce the Structural Hamming Distance between the predicted and ground-truth graphs. Because SHD has no bounded worst value, this task uses the $p_\text{worst} = p_\text{baseline}$ normalization convention of Appendix~\ref{app:metrics}.

\paragraph{Robustness (ART, dp-instahide).}
ART~\cite{nicolae2018adversarial} provides a backdoor-defense benchmark on MNIST~\cite{deng2012mnist} poisoned by five diverse attack families; the baseline is dp-instahide~\cite{borgnia2021dp}. The agent must raise the defense score, defined as the harmonic mean of clean accuracy and resistance to the five attacks.

\paragraph{Robustness (OpenOOD, MSP).}
OpenOOD~\cite{zhang2023openood} supplies a standardized OOD-detection pipeline; the in-distribution data is CIFAR-10~\cite{krizhevsky2009learning} and the OOD data is SVHN. The baseline is the Maximum Softmax Probability post-hoc scorer, and the agent must raise the near-OOD AUROC.

\paragraph{Privacy (PrivacyMeter, MIA on CIFAR-10).}
PrivacyMeter~\cite{murakonda2020ml} audits machine-learning privacy via membership inference attacks; the target model is a WRN-28-2~\cite{zagoruyko2016wide} trained on CIFAR-10~\cite{krizhevsky2009learning}. The agent must drive the attacker's AUC toward 0.5 (equivalently, reduce the MIA / RMIA AUC gap from that baseline) while preserving classification accuracy.

\paragraph{Privacy (Opacus, DP-SGD on CIFAR-10).}
Opacus~\cite{yousefpour2021opacus} trains neural networks with DP-SGD under formal $(\varepsilon, \delta)$-differential privacy; the baseline is a CNN with GroupNorm on CIFAR-10 under an $(\varepsilon{=}8, \delta{=}10^{-5})$ budget. The agent must raise test accuracy under that fixed privacy budget.

\paragraph{Fairness (AIF360, Adversarial Debiasing on COMPAS).}
COMPAS~\cite{angwin2016machine} is a recidivism-prediction dataset with a known race disparity; the baseline is Adversarial Debiasing from the AIF360~\cite{aif360-oct-2018} toolkit. The agent must reduce the absolute Average Odds Difference while maintaining classification accuracy.

\paragraph{Fairness (Fairlearn, Logistic Regression on Adult).}
Fairlearn~\cite{bird2020fairlearn} provides constrained fair classification on the UCI Adult dataset~\cite{adult_2}; the baseline is unconstrained Logistic Regression. The agent must reduce the absolute demographic-parity difference between protected and unprotected groups.

\paragraph{Machine Unlearning (Open-Unlearning, Gradient Ascent on TOFU).}
Open-Unlearning~\cite{dorna2025openunlearning} evaluates LLM unlearning on TOFU~\cite{maini2024tofu}, a fictional-author forgetting benchmark built on Llama-3.2-1B. The baseline is naive Gradient Ascent; on TOFU, the native forget-quality measure is the $p$-value of a KS test comparing the unlearned model's log-likelihoods on the forget set against a retain-only reference, and larger $p$ means the two distributions are less distinguishable (better forgetting). Because the baseline $p$-values underflow to $\approx 10^{-167}$, we apply a post-hoc display transform $-\log_{10}(p)$ for numerical stability; in this $-\log_{10}$ space the metric is ``lower is better'' (smaller $-\log_{10}(p)$ corresponds to larger $p$, hence less distinguishable outputs). The agent's objective is therefore to drive $-\log_{10}(p)$ toward $0$. Because $-\log_{10}(p)$ has no natural upper bound on the worst side, this task uses the $p_\text{worst} = p_\text{baseline}$ normalization convention of Appendix~\ref{app:metrics}.

\paragraph{Federated Learning (PFLlib, FedAvg on CIFAR-10).}
PFLlib~\cite{zhang2023pfllib} is a personalized federated-learning library; the task is CIFAR-10 split across 20 clients under non-IID data partitioning, with FedAvg~\cite{mcmahan2017fedavg} as the baseline. The agent must raise test accuracy through better aggregation or personalization.

\begin{table}[t]
\centering
\scriptsize
\setlength{\tabcolsep}{3pt}
\caption{Per-task description cards. Columns: domain, task, brief description, theoretical best ($p_{\text{best}}$), opportunity density $\phi_\text{opp}$ (mean over 6 agents $\times$ 3 rounds; Section~\ref{app:autoresearch_polarization}), partition label, and the untouched baseline test metric on its native scale. The three tasks (CausalML, gCastle, Unlearning) without a natural worst bound use the $p_{\text{worst}} = p_{\text{baseline}}$ convention defined in Appendix~\ref{app:metrics}.}
\label{tab:task_cards}
\begin{tabular}{@{}l l p{0.40\textwidth} c c c r@{}}
\toprule
\textbf{Domain} & \textbf{Task} & \textbf{Brief description} & \textbf{$p_{\text{best}}$} & \textbf{$\phi_\text{opp}$} & \textbf{Partition} & \textbf{Baseline raw} \\
\midrule
Generalization   & DomainBed-CM & Improve ERM on ColoredMNIST to maximize test-time accuracy on an unseen domain under spurious-correlation-induced distribution shift. & 1.0 & 0.118 & SPARSE-OPP & 0.287 \\
Generalization   & DomainBed-OH & Improve ERM with ResNet-50 on OfficeHome to maximize average OOD accuracy across 4 visually distinct image domains (Art, Clipart, Product, Real). & 1.0 & 0.030 & SPARSE-OPP & 0.863 \\
Data Efficiency  & EasyFSL      & Improve the Prototypical-Network few-shot classifier on Mini-ImageNet (5-way, pre-extracted ResNet-12 features) to maximize episodic accuracy. & 1.0 & 0.025 & SPARSE-OPP & 0.653 \\
Data Efficiency  & USB          & Improve FixMatch (WideResNet-28-2) on CIFAR-100 with only 200 labels plus $\sim$49.8k unlabeled samples to maximize test accuracy via pseudo-labeling and augmentation. & 1.0 & 0.047 & SPARSE-OPP & 0.079 \\
Repr.\ Learning  & Lightly      & Improve MoCo self-supervised pretraining on CIFAR-10 (no external data) to maximize frozen-encoder linear-probe top-1 accuracy. & 1.0 & 0.055 & SPARSE-OPP & 0.756 \\
Repr.\ Learning  & Solo-learn   & Improve Barlow Twins self-supervised pretraining on CIFAR-100 under a compute-constrained 200-epoch budget to maximize linear-eval accuracy. & 1.0 & 0.102 & SPARSE-OPP & 0.490 \\
Continual Learn. & Cont.-Learn. & Improve Synaptic Intelligence on class-incremental splitMNIST (no replay) to maximize average accuracy while keeping model size and training time bounded. & 1.0 & 0.254 & DENSE-OPP & 0.271 \\
Continual Learn. & PyCIL        & Improve iCaRL on CIFAR-100 class-incremental learning (50 base + 5$\times$10 new classes, 2000-exemplar memory) to maximize average incremental accuracy. & 1.0 & 0.049 & SPARSE-OPP & 0.595 \\
Causality        & CausalML     & Improve Dragonnet on IHDP to minimize mean absolute error (MAE) of individual treatment-effect estimation. & 0.0 & 0.077 & SPARSE-OPP & 1.328 \\
Causality        & gCastle      & Improve NOTEARS structure discovery on 50-node nonlinear Erd\H{o}s--R\'{e}nyi DAGs (500 samples) to minimize Structural Hamming Distance against the true graph. & 0.0 & 0.347 & DENSE-OPP & 71 \\
Robustness       & ART          & Improve the dp-instahide defense on poisoned MNIST (5 diverse backdoor / clean-label attacks) to maximize defense score while preserving clean accuracy. & 1.0 & 0.573 & DENSE-OPP & 0.477 \\
Robustness       & OpenOOD      & Improve out-of-distribution detection beyond the MSP baseline (CIFAR-10 in-dist.\ vs SVHN / other OOD sets) to maximize AUROC. & 1.0 & 0.045 & SPARSE-OPP & 0.930 \\
Privacy          & PrivacyMeter & Improve the WRN-28-2 training pipeline on CIFAR-10 to drive MIA / RMIA attacker AUC toward 0.5 (minimize AUC gap) while keeping task accuracy within baseline $\pm$1\%. & 0.0 & 0.454 & DENSE-OPP & 0.321 \\
Privacy          & Opacus       & Improve DP-SGD training of a CNN+GroupNorm on CIFAR-10 under a fixed ($\varepsilon{=}8$, $\delta{=}10^{-5}$) privacy budget to maximize test accuracy. & 1.0 & 0.144 & DENSE-OPP & 0.588 \\
Fairness         & AIF360       & Improve Adversarial Debiasing on COMPAS to minimize absolute Average Odds Difference ($|$AOD$|$) while preserving classification accuracy. & 0.0 & 0.356 & DENSE-OPP & 0.240 \\
Fairness         & Fairlearn    & Improve Logistic Regression fairness on Adult Census to minimize absolute demographic parity difference while maintaining balanced accuracy. & 0.0 & 0.447 & DENSE-OPP & 0.173 \\
Unlearning       & Unlearning   & Improve the unlearning loss (beyond Gradient Ascent) on TOFU / Llama-3.2-1B to reduce $-\log_{10}(p)$ of the forget-set KS test (raising the underlying $p$-value, i.e.\ making the unlearned outputs statistically indistinguishable from a retain-only model) while preserving utility. & $0.0$ & 1.207 & DENSE-OPP & 166.80 \\
Federated Learn. & PFLlib       & Improve FedAvg on CIFAR-10 with 20 clients under non-IID partitioning (Dirichlet $\alpha{=}0.1$) to maximize global test accuracy via better aggregation / personalization. & 1.0 & 0.352 & DENSE-OPP & 0.453 \\
\bottomrule
\end{tabular}
\end{table}

\section{Agent implementation details}\label{app:agents}

Table~\ref{tab:agent_design_long} gives the long-form description of each agent's search algorithm, with stage budgets, sampling percentages, parameter names, and implementation details.

\begin{table}[t]
\centering
\small
\renewcommand{\arraystretch}{1.15}
\caption{Detailed search algorithms of the six agents evaluated in FML-bench, with stage budgets, sampling percentages, and defining parameters.}
\label{tab:agent_design_long}
\begin{tabularx}{\textwidth}{@{}l X@{}}
\toprule
\textbf{Agent} & \textbf{Search Algorithm} \\
\midrule
The AI Scientist v1 & \textbf{Parallel Linear.} Generate a batch of \texttt{num\_ideas} independent research ideas up front (LLM reflection loop), then test each idea sequentially with up to \texttt{max\_runs} successful runs and per-run debug retries on failure (governed by a separate \texttt{max\_retries\_per\_run} counter). Ideas are \emph{parallel} in that none conditions on another's outcome (information-parallel, though executed serially here); each idea's exploration is \emph{linear}, a single chain of runs with no tree or population. \\
\midrule
The AI Scientist v2 & \textbf{Best-first tree search (BFTS)} over four stages with step-budget ratios $[0.10, 0.20, 0.50, 0.20]$, corresponding to \emph{basic implementation, hyperparameter tuning, creative research, ablation refinement}. Within each stage node selection is deterministic best-metric-first; an LLM-maintained journal that summarizes prior successful and failed experiments is injected into the code-generation prompt (the journal informs the prompt, not the selection rule). \\
\midrule
AIDE & \textbf{Solution-space tree search} with up to \texttt{num\_drafts} draft roots. At each step the agent greedily improves the best-metric \emph{good} node (any depth, not leaf-constrained); with probability \texttt{debug\_prob} the step instead expands a randomly chosen buggy leaf (any depth $\leq$ \texttt{max\_debug\_depth}) to repair it before continuing to improve. \\
\midrule
AIRA & \textbf{UCT tree search (MCTS).} Each selected node is expanded into \texttt{num\_children} children per visit; validation metrics are backpropagated without rollout. Node values are maintained as running mean fitness estimates, and observed fitness values are min--max normalized during search to keep UCT/search hyperparameters consistent across varying metric scales. \\
\midrule
Autoresearch & \textbf{Greedy hill-climbing.} After every step: keep the change if validation strictly improves, discard it otherwise, or enter a bounded debug loop on crash. The only persistent state is the current incumbent; no tree, no population, no journal. \\
\midrule
OpenEvolve & \textbf{Island-model evolutionary search with per-island MAP-Elites archive.} Each island keeps its own $5 \times 5$ feature grid (code-length $\times$ edit-distance diversity) under cell-elitist replacement. Parent selection within an island uses three strategies: \emph{exploration} (random from island, 20\%), \emph{exploitation} (from a flat global elite archive, 70\%), and \emph{weighted} (fitness-proportionate from island, 10\%); islands exchange top programs every \texttt{migration\_interval} generations via ring-topology migration, with each island donating to both neighbours. \\
\bottomrule
\end{tabularx}
\end{table}

\paragraph{The AI Scientist v1~\cite{lu2024ai}.}
The first published AI research agent of the TAS family, built around batch idea generation followed by linear per-idea execution without cross-idea feedback.

\paragraph{The AI Scientist v2~\cite{yamada2025ai}.}
A direct successor that replaces v1's linear per-idea execution with best-first tree search and adds an LLM-maintained journal to carry memory across branches.

\paragraph{AIDE~\cite{aide2025}.}
A general-purpose engineering agent whose search space is the tree of code solutions; its branching policy alternates between greedy improvement on the best good node and stochastic debugging of buggy leaves.

\paragraph{AIRA~\cite{toledo2025aira}.}
A tree-search agent that instantiates MCTS with UCT-based selection and multi-child expansion, using running mean fitness as node values and min--max normalizing observed fitness values to keep search hyperparameters consistent across varying metric scales.

\paragraph{Autoresearch~\cite{karpathy2026autoresearch}.}
A single-incumbent agent that keeps a change only when validation strictly improves, discards otherwise, and runs a bounded debug loop on crash. It is the simplest search policy in this study.

\paragraph{OpenEvolve~\cite{openevolve}.}
A population-based agent inspired by AlphaEvolve; it maintains a per-island MAP-Elites archive over a two-axis code-behavior grid, samples parents under a three-strategy mixture, and migrates top programs between islands on a ring topology.

\section{Agent porting implementation}\label{app:porting}

This appendix describes how the six agents were adapted from their original codebases to the shared FML-bench execution framework described in Section~\ref{sec:controlled_framework} and Appendix~\ref{app:codeeditor}.

\paragraph{Porting principle.}
To ensure that performance differences reflect search strategy rather than implementation artifacts, we separate each agent's codebase into three layers.
The first layer is the \emph{core search algorithm}---search topology, node selection rule, acceptance criterion, and memory mechanism---which is preserved from the original implementation and verified by line-level alignment with the source code.
The second layer is \emph{execution infrastructure}---code editing, experiment execution, metric display, and validation/test separation---which is provided by the shared FML-bench framework (Section~\ref{sec:controlled_framework}, Appendix~\ref{app:codeeditor}), replacing each agent's native execution stack.
The third layer consists of \emph{auxiliary subsystems} orthogonal to the core search decision of what to try next, which are uniformly removed across all agents.

\paragraph{Uniformly removed subsystems.}
The following modules are removed from every agent that originally included them: paper writeup and \LaTeX{} generation, automated reviewing (GPT-4o review), literature retrieval (Semantic Scholar), visual plot analysis (VLM-based), multi-seed averaging within a single step, and each agent's native code-editing tool (e.g., Aider in TAS~v1/v2, \texttt{plan\_and\_code} in AIDE). The last of these is replaced by the shared code editor of Appendix~\ref{app:codeeditor}.
These modules do not participate in the core search decision: paper writeup and reviewing modules execute only after the search is complete; multi-seed averaging refines evaluation precision rather than search direction. Literature retrieval and VLM analysis can provide supplementary input to idea generation, but they depend on external services (e.g., the Semantic Scholar API) whose availability and retrieval quality introduce uncontrolled confounds; removing them eliminates this source of confounding while ensuring cross-agent comparability.
Uniformly removing these modules may underestimate each agent's absolute performance relative to its full original configuration, but ensures that observed performance differences are attributable to search strategy.

\paragraph{Per-agent porting fidelity.}
TAS~v1 and AIDE have the highest porting fidelity: TAS~v1's parallel-linear idea chain and AIDE's solution-space tree search (with greedy best-node improvement and stochastic buggy-leaf debugging) are line-level aligned with their original implementations, with no algorithmic modifications beyond the infrastructure replacement described above.

The remaining four agents preserve their core search algorithms but involve additional framework adaptations.
TAS~v2's best-first tree search over four stages and its LLM-journal memory are fully preserved.
AIRA's UCT node selection with global fitness normalization, multi-child expansion, and backpropagation are fully preserved and line-level aligned with the original MCTS implementation.
Autoresearch's original implementation is a Claude-Code orchestration that interprets a natural-language \texttt{program.md} to coordinate the search process; we extract its search logic---greedy hill-climbing with strict-improvement acceptance and bounded debug retries---into an explicit programmatic control flow, while idea generation remains LLM-driven.
OpenEvolve's MAP-Elites island model, three-strategy parent sampling (exploration, exploitation, weighted), and ring-topology migration are fully preserved; population-size and migration-interval parameters are set so that the evolutionary dynamics (island migration, archive turnover) operate meaningfully within the shared 100-step budget.

\paragraph{Porting scope.}
The adaptations described above target cross-agent comparability of search strategies rather than faithful reproduction of each agent's original absolute performance. All infrastructure replacements and subsystem removals are applied uniformly across the six agents, with no agent-specific adaptation. All search-algorithm parameters used in the evaluation are reported in Table~\ref{tab:agent_design_long}.

\section{AdaptiveSearch agent implementation}\label{app:adaptive}

Section~\ref{sec:autoresearch_deep_dive} hypothesizes that greedy search is effective when improvement opportunities are dense, while broader exploration strategies have the advantage when opportunities are sparse, and proposes that stagnation in the validation metric can serve as an online switching signal. AdaptiveSearch is designed to test this hypothesis: it starts with greedy hill-climbing (Phase~1), monitors the validation metric online, and performs a one-shot irreversible switch to multi-branch frontier expansion (Phase~2) once it detects that the normalized improvement has stalled. Both phases execute within the same $T{=}100$ step budget. The following gives the full implementation details.

\paragraph{Phase~1: greedy hill-climbing.}
Phase~1 is algorithmically equivalent to Autoresearch (Appendix~\ref{app:agents}): it maintains a single incumbent and advances only when validation strictly improves, rolling back otherwise. In addition, Phase~1 maintains a best-so-far improvement curve: at the end of each step, it records the range-normalized cumulative improvement of the current best metric over the baseline (normalized analogously to $\hat{p}$ in Section~\ref{sec:final_metrics}). This curve is read-only with respect to the search behavior of Phase~1; its sole purpose is to provide input to the transition criterion described next.

\paragraph{Phase~1 $\to$ Phase~2 transition.}
Let $k$ denote the number of completed Phase~1 validation steps and $\mathrm{curve}[\cdot]$ the best-so-far improvement curve. The transition fires when the slope over a trailing window of $W$ steps falls below a threshold:
\begin{equation}
\mathrm{slope}_{W}[k] \;=\; \frac{\mathrm{curve}[k-1] \;-\; \mathrm{curve}[k-1-W]}{W} \;\le\; \varepsilon.
\label{eq:transition}
\end{equation}
The condition is evaluated only for $k \ge W+1$. Because the curve tracks the best-so-far value, it is monotonically non-decreasing, so the slope is always non-negative. The threshold $\varepsilon$ filters out marginal gains that the greedy rule may accept due to validation noise, ensuring that the transition fires only when the curve enters substantive stagnation. The transition is one-shot and irreversible: once fired, no mechanism exists to revert to Phase~1. A budget guard additionally requires $\mathrm{remaining} > 3$; otherwise Phase~2 is abandoned. In our implementation we set $W = 50$ and $\varepsilon = 0.0005$.

\paragraph{Phase~2: multi-branch frontier expansion.}
Upon transition, the number of branches $N$ is determined by the remaining budget ($\mathrm{remaining} \in [4,15] \Rightarrow N{=}1$;\; $[16,30] \Rightarrow N{=}2$;\; ${>}30 \Rightarrow N{=}3$). Each branch forks from one of the top-$N$ highest-performing codebases produced during Phase~1; if Phase~1 yields fewer than $N$ candidates, the deficit branches fork from the baseline. Branches are scheduled in strict round-robin, with each step advancing exactly one branch. The acceptance rule within each branch is identical to Phase~1 but operates on the branch-local incumbent.

Each branch's first idea is required to differ from all sibling branches at the mechanism level---distinct algorithm families, mathematical formulations, or conceptual paths; hyperparameter adjustments alone do not qualify---ensuring qualitative separation of the initial exploration directions.

\paragraph{Adaptive search guidance.}
After the first step, the agent diagnoses the branch-local search geometry before generating each subsequent idea and steers exploration accordingly. Two complementary conditions capture two distinct stagnation patterns:
\begin{itemize}[leftmargin=*,nosep]
    \item \textbf{Shallow stagnation}: fewer than $\kappa_a$ improvements in the last $W_a$ steps and $D_{\mathrm{reach}}$ below a threshold~$\tau_r$. The search has stalled without leaving the baseline neighborhood; the agent steers exploration toward deeper algorithmic variants.
    \item \textbf{Divergent stagnation}: fewer than $\kappa_b$ improvements in the last $W_b$ steps and $D_{\mathrm{eff}}$ above a threshold~$\tau_d$. The search has stalled while scattering across multiple directions; the agent steers exploration back toward refinement within the most recently successful direction.
    \item When both conditions hold simultaneously, the agent steers toward consolidation within the most promising direction combined with deeper algorithmic modification.
\end{itemize}
$D_{\mathrm{reach}}$ and $D_{\mathrm{eff}}$ are computed on the branch-local embedding sequence with definitions identical to Section~\ref{sec:metrics}. Diagnosis is suppressed when the branch history contains fewer than $\min(W_a, W_b)$ steps to avoid spurious triggering from insufficient samples. In our implementation, $D_{\mathrm{reach}}$ is normalized by a per-task calibration constant $C_{\mathrm{task}}$ (the maximum exploration reach observed across the 324 main-experiment runs for that task); thresholds are set to $W_a{=}20$, $\kappa_a{=}2$, $\tau_r{=}0.30$, $W_b{=}5$, $\kappa_b{=}2$, $\tau_d{=}1.25$.

\section{Full metric definitions}\label{app:metrics}

This section gives the complete formal specification of all twelve process-level metrics, including edge-case conventions, together with the normalized test improvement used as the outcome metric throughout the paper. Section~\ref{sec:metrics} introduces each metric with summary definitions; the definitions below are the authoritative reference and additionally cover the unbounded-worst normalization convention, the pairwise win-rate formula, and boundary conditions for undefined cells.
Consider an agent conducting research for $T = 100$ optimization steps on a fixed task. At each valid step $t \in \{1, \ldots, n\}$ (where $n \leq T$ is the number of steps that produced a non-null result) the agent produces a codebase $\mathbf{C}_t$ with GraphCodeBERT~\cite{guo2020graphcodebert} embedding $\mathbf{e}_t$ and validation metric $\text{perf}_\text{val}(\mathbf{C}_t)$. Let $p_\text{agent}$ denote the test metric of the best-validated codebase selected at the end of the run, let $p_\text{baseline}$ denote the untouched baseline metric, and let $p_\text{best}/p_\text{worst}$ denote the theoretical best and worst of the task metric (e.g.\ $1.0/0.0$ for accuracy, $0.0/1.0$ for $|$AOD$|$, and $0.0/0.5$ for the PrivacyMeter AUC gap; for the three unbounded-worst tasks we use the convention defined in the next paragraph).

\paragraph{Normalized test improvement.}
We restate and extend the definitions from Section~\ref{sec:final_metrics} with full technical detail.
The primary outcome metric, denoted $\hat{p}$. For ``higher is better'' tasks,
\[
\hat{p} = \max\!\left(0,\; \frac{p_\text{agent} - p_\text{baseline}}{|p_\text{best} - p_\text{worst}|}\right);
\]
for ``lower is better'' metrics we negate the numerator,
\[
\hat{p} = \max\!\left(0,\; \frac{p_\text{baseline} - p_\text{agent}}{|p_\text{best} - p_\text{worst}|}\right).
\]
Improvements below the baseline are set to $0$, and the denominator places the 15 bounded tasks on a common $[0, 1]$ scale. Whenever the paper refers to the \emph{normalized test improvement} (or \emph{normalized validation improvement}, computed by replacing $p_\text{agent}$ with the agent's validation metric), it refers to $\hat{p}$ as defined here.

\paragraph{Unbounded-worst convention.}
Three of the 18 tasks (CausalML, gCastle, Unlearning) use primary metrics whose worst value is not naturally bounded: the individual-treatment-effect MAE on IHDP (CausalML), the Structural Hamming Distance on 50-node DAGs (gCastle), and the $-\log_{10}(p)$ forget-quality score on TOFU (Unlearning). For these we set $p_\text{worst} = p_\text{baseline}$ in the normalization formula above, which yields a normalized improvement that is non-negative and unbounded above but still comparable within task. The 18-task MEAN row of Table~\ref{tab:normalized_performance} therefore mixes per-task scales for these three rows and should be read with that caveat.

\paragraph{Choice of code embedding model.}
GraphCodeBERT is pretrained with a data-flow graph objective (edge prediction and variable alignment) alongside masked language modeling, so its embeddings tend to reflect data-flow dependencies rather than surface token changes alone---a property favorable for this study, where agents' modifications often involve variable names or formatting but the metrics aim to capture algorithmic-level structural change. Indirectly, the four embedding-based exploration metrics exhibit selectively differentiated correlation patterns (Reach positively correlated, Effective dim negatively correlated, Uniqueness uncorrelated), broadly consistent with the qualitative analysis of Section~\ref{sec:alignment}; this suggests the embeddings more likely capture meaningful structural variation than noise, though we have not conducted a formal cross-embedding-model consistency check.

\paragraph{Exploration Spread.}
The mean L2 distance of per-step code embeddings from their centroid,
\begin{equation}
D_\text{spread} = \frac{1}{n} \sum_{t=1}^{n} \left\| \mathbf{e}_t - \bar{\mathbf{e}} \right\|_2,
\qquad
\bar{\mathbf{e}} = \frac{1}{n} \sum_{t=1}^{n} \mathbf{e}_t,
\label{eq:spread_app}
\end{equation}
where $\mathbf{e}_t$ is the GraphCodeBERT embedding of the full concatenated target-file source at the $t$-th valid step, obtained from the model's \texttt{[CLS]} token summed across fixed-length chunks. GraphCodeBERT is pretrained with a data-flow-graph objective alongside masked language modeling~\cite{guo2020graphcodebert}, so its embeddings carry data-flow structure in addition to surface tokens; concretely, two code snippets that only rename variables but preserve the same computation graph are mapped to nearly identical vectors, while two snippets with different data-flow structure but similar token soup are separated. $D_\text{spread}$ therefore measures the mean within-trajectory dispersion in this structural program space rather than surface-token churn: larger $D_\text{spread}$ means the agent's attempts travel farther from their running centroid in a sense that tracks algorithmic, not lexical, change.

\paragraph{Exploration Uniqueness.}
The cardinality counterpart to spread. Using the same embeddings $\{\mathbf{e}_t\}_{t=1}^{n}$, we run cosine agglomerative clustering with average linkage and threshold $\tau = 0.015$ and define
\[
U = \frac{\#\text{clusters}(\{\mathbf{e}_t\}_{t=1}^{n}; \tau = 0.015)}{n}.
\]
Uniqueness captures how many distinct solution families the agent's attempts fall into, independent of how far apart they lie in embedding space.

\paragraph{Exploration Reach.}
The maximum L2 distance of any per-step embedding from the baseline embedding,
\begin{equation}
D_\text{reach} = \max_{t \in \{1,\ldots,n\}} \left\| \mathbf{e}_t - \mathbf{e}_\text{base} \right\|_2,
\label{eq:reach_app}
\end{equation}
where $\mathbf{e}_\text{base}$ is the GraphCodeBERT embedding of the unmodified baseline codebase. Whereas $D_\text{spread}$ measures the trajectory's mean dispersion around its own running centroid, $D_\text{reach}$ records the furthest point the trajectory ever reaches from the starting baseline in the same structural program space.

\paragraph{Effective dim.}
The participation ratio of the centered embedding cloud,
\begin{equation}
D_\text{eff} = \frac{\left( \sum_i \lambda_i \right)^2}{\sum_i \lambda_i^2},
\label{eq:effdim_app}
\end{equation}
where $\{\lambda_i\}$ are the eigenvalues of the sample covariance $\frac{1}{n-1}\sum_{t=1}^{n} (\mathbf{e}_t - \bar{\mathbf{e}})(\mathbf{e}_t - \bar{\mathbf{e}})^\top$. $D_\text{eff}$ is small when the trajectory concentrates along a few principal directions ($\approx 1$ for a single dominant direction) and large when it spreads isotropically across many directions; on the pooled 324-run sample, low $D_\text{eff}$ correlates positively with test improvement (Section~\ref{sec:correlation_analysis}).

\paragraph{Val-test gap.}
Two forms are used. The signed val-test gap is $(\hat{p}_\text{val} - \hat{p}_\text{test})$, where $\hat{p}_\text{val}$ and $\hat{p}_\text{test}$ are the normalized improvements on the validation and test splits of the best-validated codebase. Positive signed gaps indicate overfit to the validation feedback; negative signed gaps indicate that the test split is harder than the validation split. The absolute val-test gap is the unsigned form, $|\hat{p}_\text{val} - \hat{p}_\text{test}|$; this is the metric reported in Table~\ref{tab:process_metrics}.

\paragraph{Valid step ratio.}
$R_\text{valid} = n / T$, the fraction of the $T = 100$ optimization steps that produced a valid result, i.e., neither an execution error, a timeout, nor a task-specific constraint violation.

\paragraph{AUC-over-steps.}
The time-averaged best-so-far validation improvement,
\[
\text{AUC} = \frac{1}{T} \sum_{t=1}^{T} P_\text{best}^{(t)},
\qquad
P_\text{best}^{(t)} = \max_{s \leq t} \hat{p}_\text{val}(\mathbf{C}_s),
\]
where $\hat{p}_\text{val}(\mathbf{C}_s)$ is the normalized validation improvement of the codebase at step $s$. AUC captures \emph{when} gains are realized over the step budget: agents that improve earlier and hold their lead have higher AUC.

\paragraph{First-improvement step.}
The first step $t$ at which $\hat{p}_\text{val}(\mathbf{C}_t) > 0$, i.e., the agent's validation metric first surpasses the baseline. Left undefined (NaN) if no step improves over the baseline; this explains why the pooled correlation for this metric uses $n = 305$ rather than 324.

\paragraph{Best-improvement step.}
The step $t^{\star} \in \{1, \ldots, T\}$ at which the best validation metric in the run was first achieved. On \sparse{} tasks, where marginal gains are scarce, this metric tends to be larger (late) for the top-performing agents (Section~\ref{sec:correlation_analysis}).

\paragraph{Late-gain fraction.}
The fraction of the agent's final normalized improvement that is earned in the \emph{second half} of the step budget,
\[
\text{late\_gain} = \frac{P_\text{best}^{(T)} - P_\text{best}^{(T/2)}}{P_\text{best}^{(T)}},
\]
using $T = 100$ and $T/2 = 50$. Higher values indicate back-loaded progress; lower values indicate front-loaded progress. Cells where $P_\text{best}^{(T)} = 0$ are left as NaN.

\paragraph{Token cost.}
The total number of prompt and completion tokens across all LLM calls during the run, including code-editing calls, idea-generation calls, and journal maintenance.

\paragraph{Wall-clock time.}
The total run duration in hours, measured from the first agent action to the final end-of-run test evaluation.

\paragraph{Pairwise head-to-head win-rate.}
For each ordered pair of agents $(A, B)$ and each task $t$, agent $A$ \emph{strictly wins} task $t$ if its raw test metric is strictly better than $B$'s in the task's native direction; ties are not counted as wins. Agent $A$'s overall win-rate is
\[
\mathrm{WR}(A) = \frac{1}{(K-1)\,N} \sum_{\substack{B \in \mathcal{A} \\ B \neq A}} \sum_{t=1}^{N} \mathbf{1}\!\left[A \succ_t B\right],
\]
where $K = |\mathcal{A}|$ is the number of agents, $N$ is the number of tasks, and $A \succ_t B$ denotes that $A$'s raw test metric is strictly better than $B$'s on task $t$ in that task's native direction. The denominator $(K-1)\,N$ is the maximum number of strict wins an agent can accumulate, so $\mathrm{WR}(A) \in [0, 1]$. This metric is rank-based and invariant to monotone rescaling of the test metric; it is the normalization-invariant complement to the mean normalized improvement of Table~\ref{tab:normalized_performance}.

\section{Process metrics by research-task difficulty}\label{app:process_splits}

\begin{figure}[t]
\centering
\includegraphics[width=0.95\linewidth]{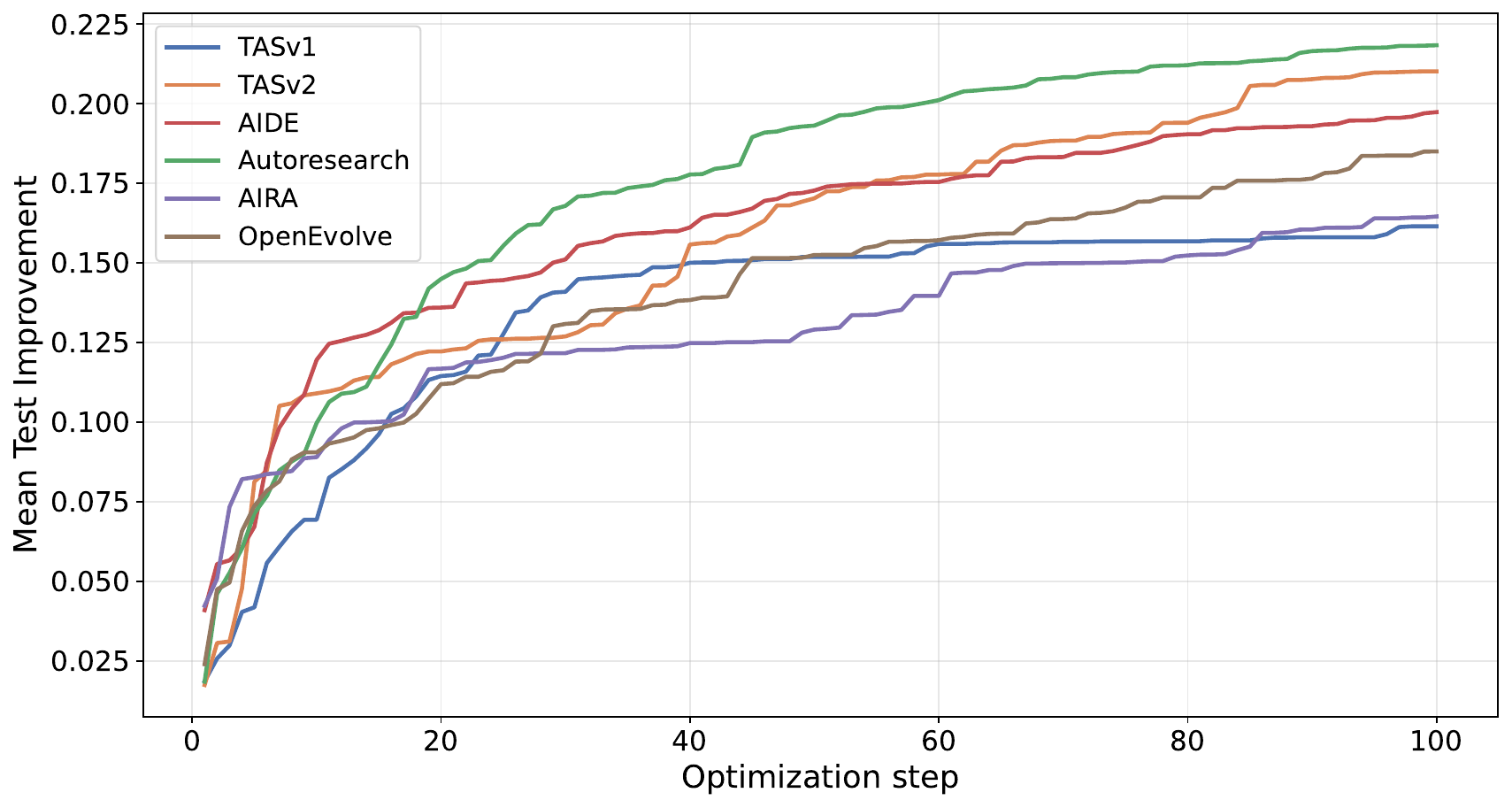}
\caption{\textbf{Mean convergence curves across 18 research tasks.} Each line is per-agent mean best-so-far validation improvement, averaged over 18 tasks $\times$ 3 rounds, at each of the 100 optimization steps.}
\label{fig:auc_convergence}
\end{figure}

This appendix opens with the per-agent mean convergence curves across all 18 tasks (Figure~\ref{fig:auc_convergence}), and then separates the dense / sparse partitions. The pooled per-agent process metrics and Spearman correlations of Table~\ref{tab:process_metrics} collapse the 9 \dense{} and 9 \sparse{} tasks defined in Section~\ref{app:autoresearch_polarization}. Table~\ref{tab:process_metrics_dense_opp} restricts the analysis to the \dense{} tasks; Table~\ref{tab:process_metrics_sparse_opp} restricts it to the \sparse{} tasks. Per-agent means are computed on the 9 tasks of each partition; $\rho$ and $p$ are pooled Spearman correlations over the 162 (round, agent, task) tuples of the same partition. Both tables mirror the structure of Table~\ref{tab:process_metrics}.

\begin{table}[t]
\centering
\footnotesize
\setlength{\tabcolsep}{3pt}
\caption{Process metrics restricted to the 9 \dense{} tasks (162 (round, agent, task) tuples). AUC-over-steps and First-improvement step are the dominant signals; Wall-clock time and Valid step ratio also reach significance. Significance: *** $p < 0.001$, ** $p < 0.01$, * $p < 0.05$.}
\label{tab:process_metrics_dense_opp}
\begin{tabular}{llrrrrrrrr}
\toprule
Dimension & Metric & TAS v1 & TAS v2 & AIDE & AIRA & AutoR & OEvolve & $\rho$ & $p$ \\
\midrule
Exploration & Exploration Spread & 24.50 & 30.91 & 16.20 & 15.79 & 19.99 & 11.24 & +0.081 & 0.305 \\
Exploration & Exploration Uniqueness & 0.0733 & 0.0733 & 0.0519 & 0.0437 & 0.0437 & 0.1337 & $-$0.050 & 0.531 \\
Exploration & Exploration Reach & 141.5 & 154.2 & 74.66 & 81.36 & 110.2 & 49.25 & +0.043 & 0.586 \\
Exploration & Effective dim & 1.633 & 1.416 & 2.926 & 2.315 & 1.406 & 2.845 & $-$0.122 & 0.122 \\
Generalization & Val-test $|$gap$|$ & 0.0584 & 0.0311 & 0.0394 & 0.0728 & 0.0440 & 0.0759 & +0.053 & 0.501 \\
Reliability & Valid step ratio & 0.7937 & 0.7830 & 0.8733 & 0.8574 & 0.8270 & 0.7733 & +0.160 * & 0.042 \\
Efficiency & AUC-over-steps & 0.2411 & 0.2752 & 0.2790 & 0.2329 & 0.3134 & 0.2317 & \textbf{+0.830} *** & 2.5e-42 \\
Efficiency & First-improvement step & 9.958 & 10.81 & 9.463 & 9.667 & 7.704 & 17.15 & \textbf{$-$0.437} *** & 3.1e-08 \\
Efficiency & Late-gain fraction & 0.0731 & 0.1983 & 0.0533 & 0.2482 & 0.0856 & 0.2523 & $-$0.018 & 0.833 \\
Efficiency & Best-improvement step & 49.07 & 71.37 & 79.48 & 75.59 & 94.81 & 78.81 & $-$0.100 & 0.203 \\
Cost & Token cost (M) & 2.008 & 1.269 & 1.533 & 0.9235 & 1.364 & 0.7387 & +0.087 & 0.269 \\
Cost & Wall-clock time (h) & 21.73 & 23.71 & 17.96 & 20.05 & 22.53 & 19.69 & +0.171 * & 0.029 \\
\bottomrule
\end{tabular}
\end{table}

\begin{table}[t]
\centering
\footnotesize
\setlength{\tabcolsep}{3pt}
\caption{Process metrics restricted to the 9 \sparse{} tasks (162 (round, agent, task) tuples). Three metrics null on the pooled sample become significant here: Late-gain fraction, Best-improvement step, and Val-test $|$gap$|$ (which flips sign). Significance: *** $p < 0.001$, ** $p < 0.01$, * $p < 0.05$.}
\label{tab:process_metrics_sparse_opp}
\begin{tabular}{llrrrrrrrr}
\toprule
Dimension & Metric & TAS v1 & TAS v2 & AIDE & AIRA & AutoR & OEvolve & $\rho$ & $p$ \\
\midrule
Exploration & Exploration Spread & 22.81 & 25.98 & 16.25 & 13.42 & 26.70 & 10.54 & +0.077 & 0.330 \\
Exploration & Exploration Uniqueness & 0.0522 & 0.0215 & 0.0196 & 0.0374 & 0.0244 & 0.0256 & +0.026 & 0.747 \\
Exploration & Exploration Reach & 103.7 & 124.9 & 71.28 & 61.02 & 111.1 & 35.54 & +0.026 & 0.744 \\
Exploration & Effective dim & 1.934 & 1.483 & 3.022 & 3.174 & 2.012 & 4.591 & $-$0.065 & 0.413 \\
Generalization & Val-test $|$gap$|$ & 0.0139 & 0.0183 & 0.0153 & 0.0074 & 0.0229 & 0.0138 & \textbf{$-$0.180} * & 0.022 \\
Reliability & Valid step ratio & 0.7552 & 0.6848 & 0.9041 & 0.7211 & 0.8211 & 0.8830 & +0.007 & 0.928 \\
Efficiency & AUC-over-steps & 0.0301 & 0.0405 & 0.0434 & 0.0277 & 0.0376 & 0.0508 & \textbf{+0.589} *** & 1.6e-16 \\
Efficiency & First-improvement step & 15.30 & 21.48 & 11.78 & 21.46 & 10.67 & 13.07 & +0.017 & 0.832 \\
Efficiency & Late-gain fraction & 0.1365 & 0.3933 & 0.1837 & 0.3928 & 0.2907 & 0.1969 & \textbf{+0.242} ** & 0.002 \\
Efficiency & Best-improvement step & 48.33 & 76.96 & 78.70 & 67.96 & 87.67 & 63.07 & \textbf{+0.184} * & 0.019 \\
Cost & Token cost (M) & 2.568 & 1.583 & 1.942 & 1.226 & 2.108 & 1.130 & $-$0.067 & 0.396 \\
Cost & Wall-clock time (h) & 38.81 & 41.27 & 36.75 & 38.11 & 36.28 & 37.71 & +0.072 & 0.361 \\
\bottomrule
\end{tabular}
\end{table}

\section{Opportunity density, task partition, and Autoresearch polarization detail}\label{app:autoresearch_polarization}
 
Section~\ref{sec:autoresearch_deep_dive} hypothesizes that the effectiveness of a search strategy depends on a task's improvement opportunity structure. This appendix defines a post-hoc metric to quantify this structure, partitions the 18 tasks accordingly, and presents supporting evidence for the hypothesis.
 
\paragraph{Opportunity density.}
For each (round, agent, task) cell with $n$ valid steps, let $\mathrm{imp}_k$ denote the normalized improvement at step $k$ (Section~\ref{sec:final_metrics}), $d_k = \lVert \mathbf{e}_k - \mathbf{e}_{\mathrm{base}} \rVert_2$ the GraphCodeBERT embedding distance from the baseline, and $d_{\max} = \max_{1 \le k \le n} d_k$ the exploration reach. The \emph{opportunity density} of a cell is
\begin{equation}
\phi_{\text{opp}} \;=\; \frac{\displaystyle\sum_{k \in \mathcal{P}} \mathrm{imp}_k}{\displaystyle\sum_{k \in \mathcal{P}} d_k \,/\, d_{\max}}, \quad \mathcal{P} = \{k : \mathrm{imp}_k > 0\},\ \ d_{\max} = \max_{1 \le k \le n} d_k.
\label{eq:opp_density}
\end{equation}
Higher $\phi_{\text{opp}}$ indicates that gains accumulate quickly per unit embedding distance travelled; lower $\phi_{\text{opp}}$ indicates that gains are sparse. The per-task value $\phi_{\text{opp}}(t)$ is the mean over 6 agents $\times$ 3 rounds. Because $\phi_{\text{opp}}$ is measured from the agent runs rather than from baseline-only quantities, the resulting partition reflects what agents actually find rather than what is a priori available above the baseline.
 
\paragraph{Task partition.}
Splitting the 18 tasks at the median value $\phi_{\text{opp}} = 0.1315$ yields two groups: \dense{} (top nine tasks, e.g.\ Unlearning, ART, PrivacyMeter) and \sparse{} (bottom nine tasks, e.g.\ DomainBed-OH, OpenOOD, EasyFSL). Per-task $\phi_{\text{opp}}$ values and partition assignments are listed in Table~\ref{tab:task_cards}.
 
\paragraph{Partition evidence.}

\begin{figure}[t]
\centering
\includegraphics[width=0.95\linewidth]{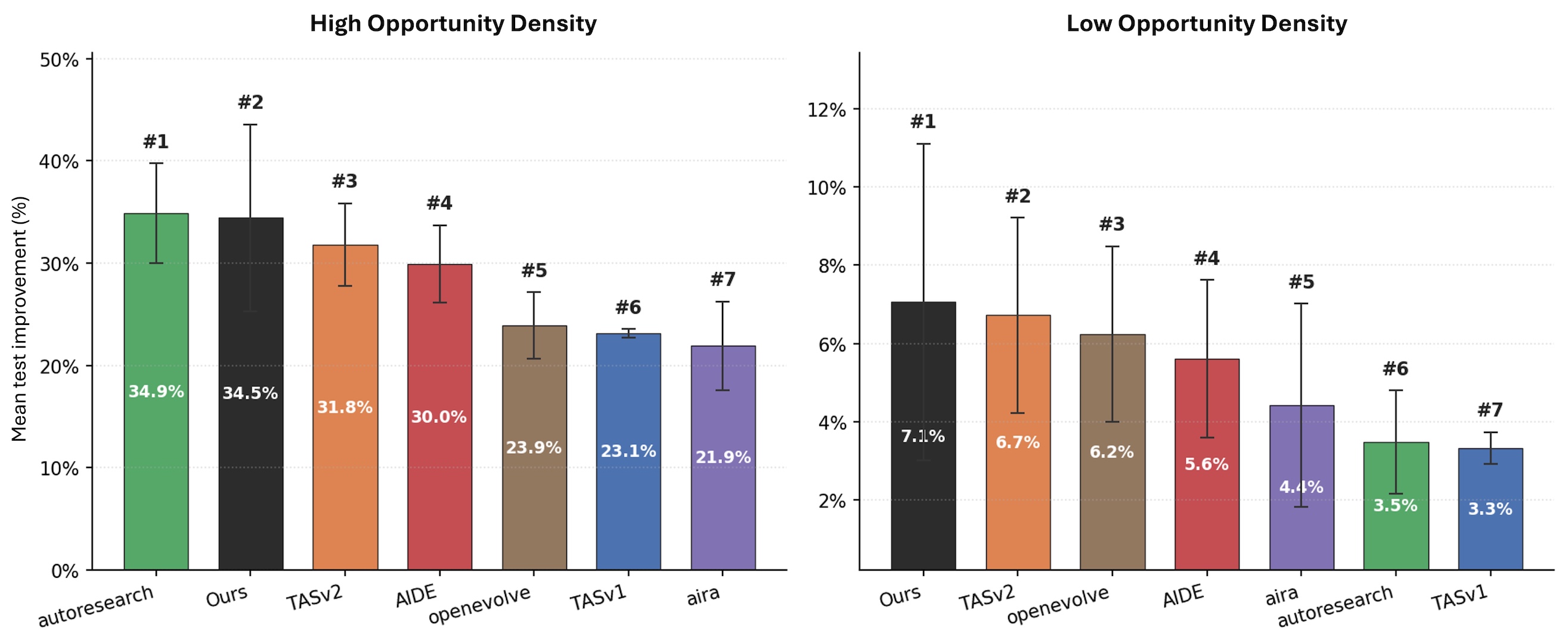}
\caption{\textbf{Search-regime crossover.} Per-agent mean normalized test improvement on the high and low opportunity-density partitions; error bars are the cross-round standard deviation. Autoresearch leads the high-density partition but falls to sixth (of seven) on low-density; AdaptiveSearch ranks in the top two on both partitions (second on high-density, first on low-density), confirming that adaptive switching is robust across both task regimes.}
\label{fig:partition_mean_bars}
\end{figure}

Figure~\ref{fig:partition_mean_bars} shows the ranking inversion this partition exposes. On the nine \dense{} tasks, Autoresearch posts the highest mean test improvement ($0.349$), with AdaptiveSearch close behind ($0.345$); on the nine \sparse{} tasks, Autoresearch drops to sixth among seven agents ($0.035$), while AdaptiveSearch leads ($0.071$), followed by TAS~v2 ($0.067$) and OpenEvolve ($0.062$). Autoresearch's cross-partition drop from first to sixth remains the largest ranking inversion of any agent in the suite, providing post-hoc evidence for the hypothesis of Section~\ref{sec:autoresearch_deep_dive}. AdaptiveSearch is the only agent that ranks in the top two on both partitions, confirming that adaptive switching covers both task regimes.

\begin{figure}[t]
\centering
\includegraphics[width=0.95\linewidth]{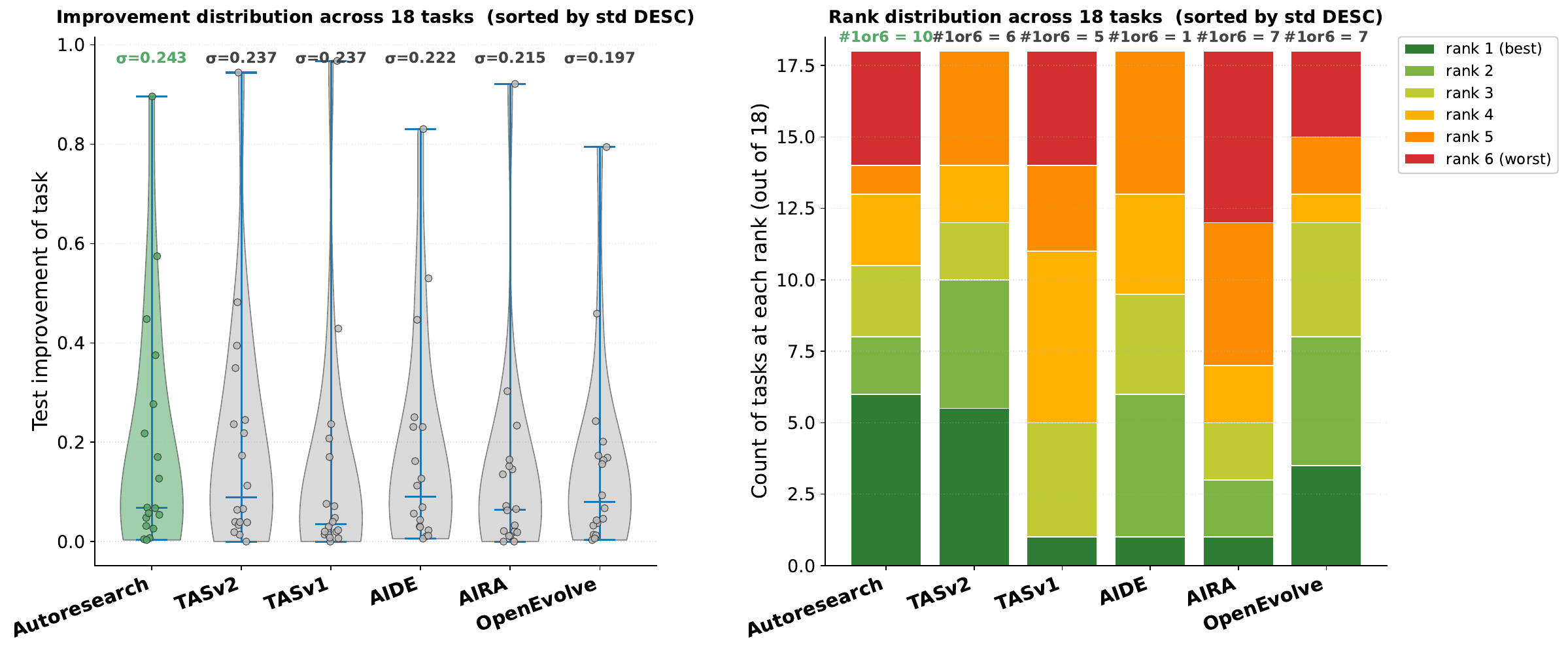}
\caption{\textbf{Autoresearch's per-task improvement is the most polarized of the six agents.} \emph{Left:} per-agent improvement distribution across the 18 tasks (3-round mean per cell), agents sorted by std. \emph{Right:} per-task rank distribution (rank 1 best, rank 6 worst). Autoresearch attains the largest improvement std and the most extreme rank distribution.}
\label{fig:agent_polarization}
\end{figure}

\paragraph{Autoresearch polarization.}
Autoresearch's per-task improvement standard deviation ($0.243$) is the largest of the six agents, and ten of its eighteen tasks land at either rank~1 or rank~6, the most polarized rank distribution in the suite. Figure~\ref{fig:agent_polarization} visualizes the two distributions backing that claim.
 
\section{Backbone LLMs: Quality, Efficiency, and Search Behavior}
\label{sec:llm-backbones}

We isolate the effect of the backbone LLM by fixing the agent to Autoresearch and varying only the underlying model. All runs use the same 18 FML-bench tasks, the same 100-step budget, and the same range-normalized non-negative test-improvement metric. Under this controlled setting, no single backbone dominates every operating objective: Gemini 3.1 Pro provides the strongest final quality, GPT-5.4 is the most cost-efficient, and Claude Opus 4.6 converges fastest.

\begin{figure}[t]
  \centering
  \includegraphics[width=0.48\linewidth]{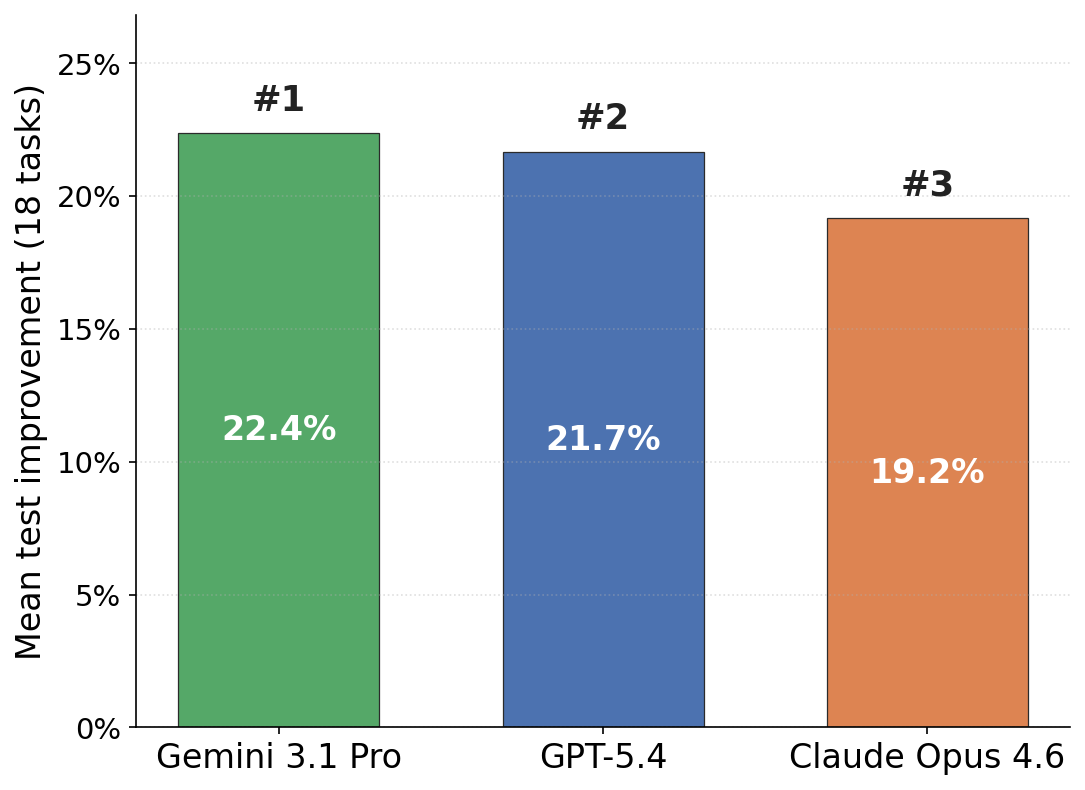}\hfill
  \includegraphics[width=0.48\linewidth]{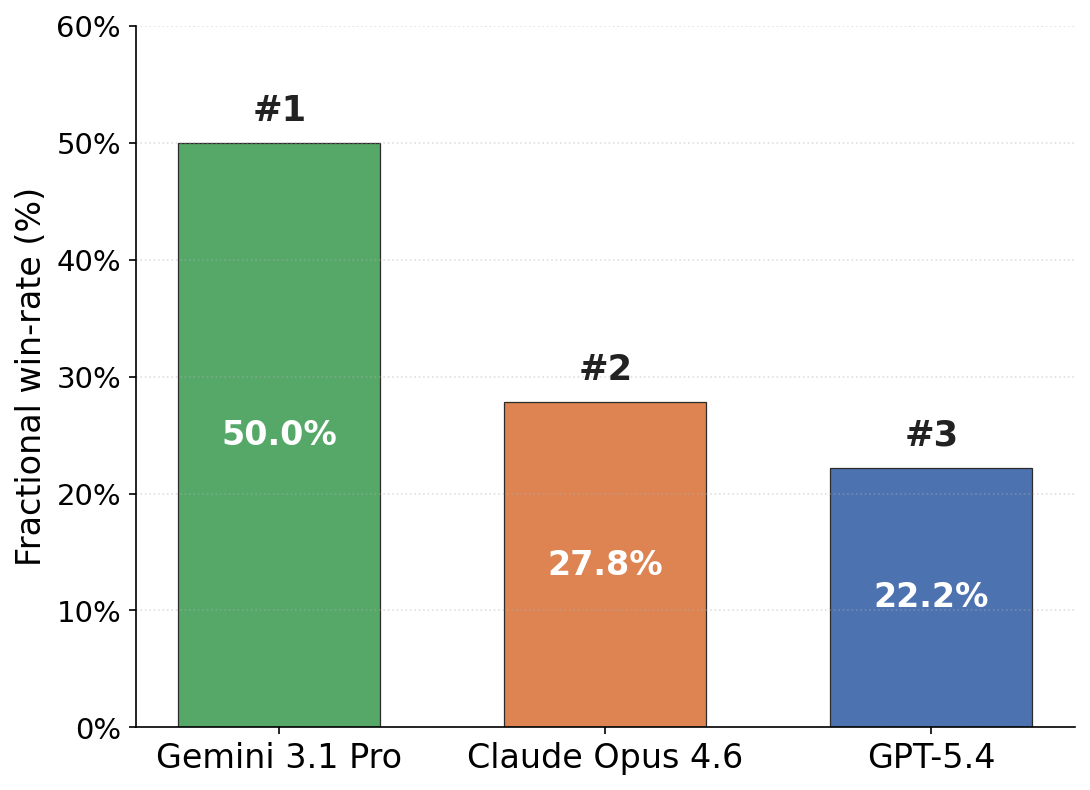}
  \caption{Raw quality comparison across backbone LLMs. Gemini 3.1 Pro is the most consistent backbone, achieving the strongest aggregate improvement and the highest task win rate. GPT-5.4 remains competitive in mean improvement, while Claude Opus 4.6 provides a complementary but weaker full-budget profile.}
  \label{fig:llm-quality-summary}
\end{figure}

Figure~\ref{fig:llm-quality-summary} summarizes the main quality results. Gemini 3.1 Pro achieves the highest mean test improvement, the highest median test improvement, and the largest number of task wins. This indicates that its advantage comes from consistent performance across tasks rather than a single outlier. GPT-5.4 remains close in mean improvement, but its much lower median suggests a more heavy-tailed profile, with performance driven by occasional large gains. Claude Opus 4.6 trails Gemini 3.1 Pro in final quality, but remains competitive on typical-task performance and reaches its best solutions earlier.

\begin{figure}[t]
  \centering
  \includegraphics[width=0.7\linewidth]{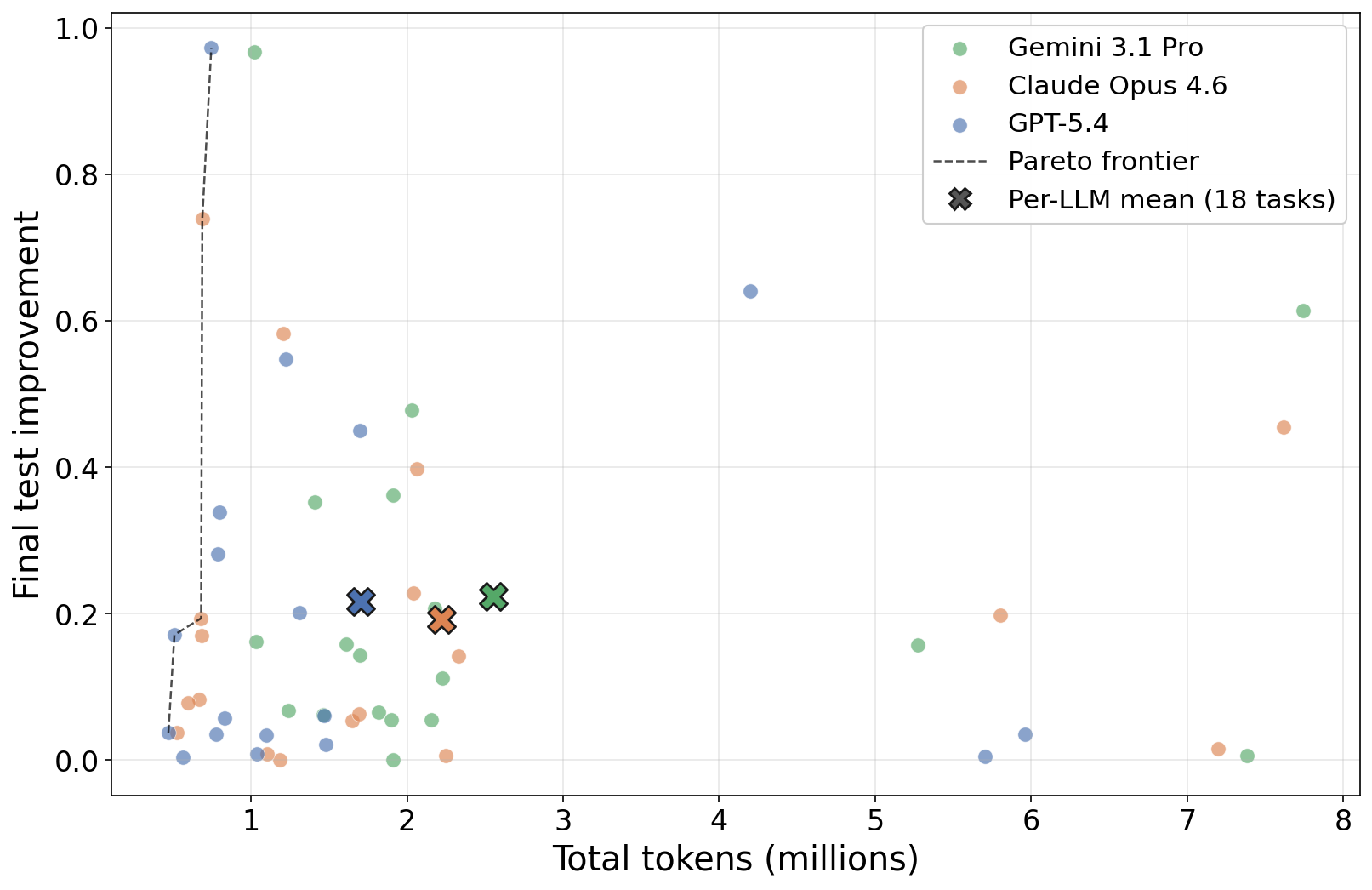}
  \caption{Cost--quality trade-off across backbone LLMs. GPT-5.4 occupies the low-cost regime while maintaining competitive quality, making it the strongest choice when optimization efficiency is the primary concern.}
  \label{fig:cost-performance}
\end{figure}

These differences reflect distinct optimization regimes. Gemini 3.1 Pro is the strongest default for maximum final quality: it has the highest valid-step ratio and reliably produces useful candidate programs within the same step budget. GPT-5.4 is the most budget-efficient backbone: it uses the fewest tokens while staying competitive in final quality, but its gains arrive late, so its value is most apparent under the full 100-step budget. Claude Opus 4.6 converges fastest, making it useful for pilot runs or early directional feedback, but its early saturation does not translate into the best full-budget quality. Figure~\ref{fig:cost-performance} shows the corresponding cost--quality trade-off.

Two negative findings are important. First, no simple search-behavior statistic alone explains final quality across backbone LLMs. Second, higher token usage does not explain quality differences. The main conclusion is therefore not that a single process metric predicts success, but that different backbone LLMs induce different optimization regimes, with distinct trade-offs in reliability, efficiency, and convergence speed.

\section{Modification and Failure Distribution}
\label{sec:agent_behavior_signatures}

We next ask whether different agent strategies leave measurable behavioral signatures in FML-bench. We focus on two axes: the \emph{modification distribution}, which records what type of intervention an agent proposes, and the \emph{failure distribution}, which records why a trial does not complete successfully. This distinction is important because failures can arise from invalid code, wall-clock timeouts, or trials whose final metric is invalid after execution.

All error and modification types are manually annotated by one PhD student. Failure types are annotated for every failed step across all agent--task--run combinations, covering $6$ agents, $18$ tasks, $3$ runs, and up to $100$ steps per run. Modification types are annotated on a sampled subset: for each agent--task--run combination, we randomly sample $5$ steps from the $100$-step trajectory, yielding $6 \times 18 \times 3 \times 5$ annotated modification steps in total.

\begin{figure}[t]
    \centering
    \includegraphics[width=0.98\linewidth]{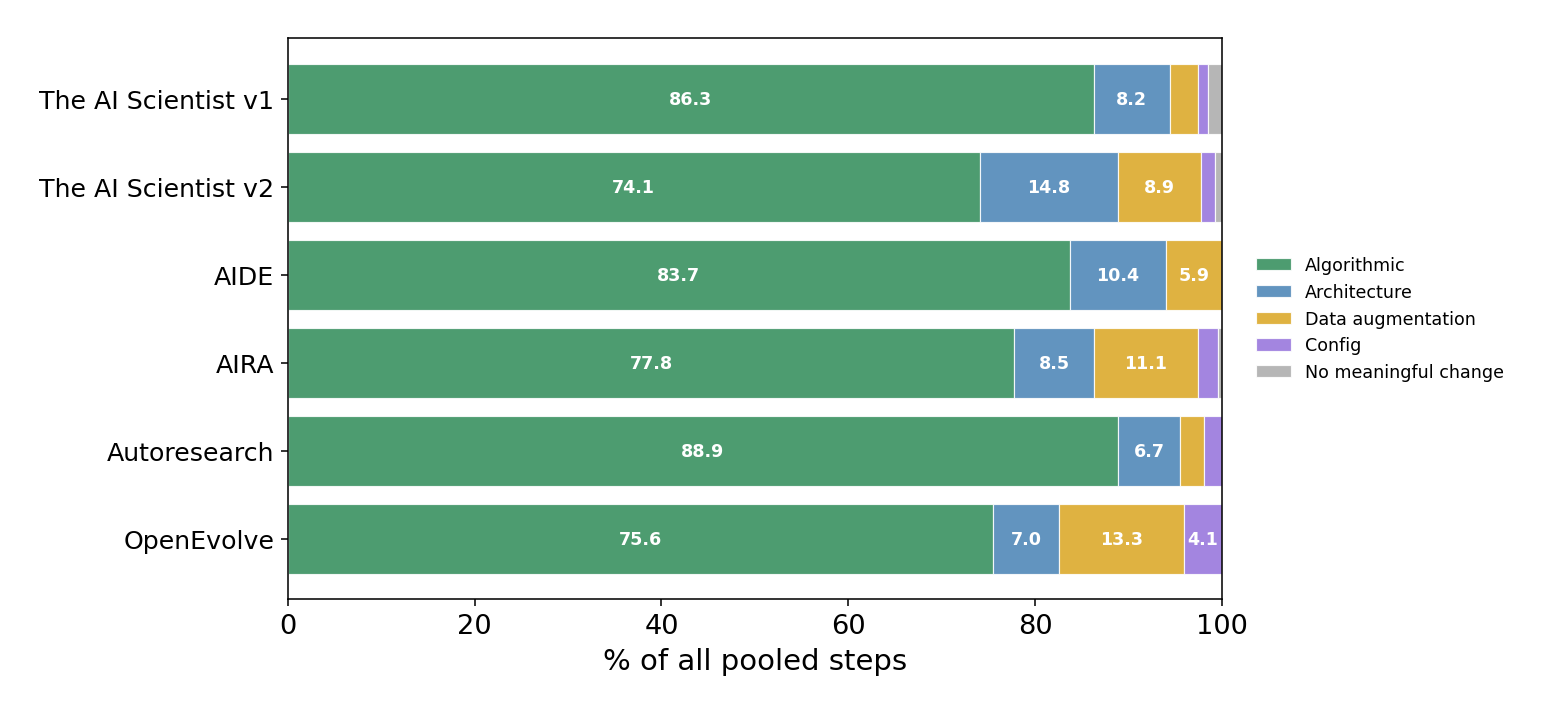}
    \caption{
    Pooled modification-type distribution across three runs for each agent. All agents are algorithmic-first, but differ in how they allocate the remaining edits to architecture, data augmentation, configuration, and no meaningful change.
    }
    \label{fig:modification_pct_pooled}
\end{figure}

Figure~\ref{fig:modification_pct_pooled} shows the pooled modification-type distribution across three runs. The dominant pattern is that all agents are algorithmic-first: most proposed edits modify the learning recipe, including losses, optimizers, schedules, sampling strategies, or update rules. This holds across greedy hill-climbing, best-first tree search, MCTS, solution-space search, and evolutionary search. The informative differences therefore lie in the remaining probability mass: whether an agent spends its non-algorithmic budget on architecture, data-pipeline, configuration, or no-op changes.

The modification profiles are consistent with the search state maintained by each agent. Autoresearch and The AI Scientist v1 are closest to recipe tuners. Autoresearch keeps only a single incumbent and accepts edits only when validation improves, so local algorithmic changes are the safest way to obtain immediate gains. The AI Scientist v1 also favors algorithmic edits because each generated idea is evaluated as an independent linear chain, with no tree or population to preserve broader structural variants. AIDE remains mostly algorithmic, but its solution-space tree and explicit buggy-leaf repair allow slightly broader yet bounded changes: the agent can improve promising nodes while still recovering from failed branches. The AI Scientist v2 is the main structural-edit outlier. Its staged best-first search reserves a large portion of the budget for creative research and ablation refinement, and its journal provides context from previous experiments; together, these mechanisms encourage larger architecture-level interventions. AIRA and OpenEvolve allocate more mass to data-side changes because their designs explicitly preserve exploration. AIRA expands multiple children from selected nodes under UCT, repeatedly exposing the search to alternative sampling or augmentation choices, while OpenEvolve's island-model MAP-Elites archive preserves diverse programs as separate niches, allowing data-pipeline variants to survive even when they are not globally best.

However, modification type alone does not predict performance or reliability. A high algorithmic-edit share is not automatically safe, and broader structural editing is not always harmful. The effect of an edit type depends on both the agent and the task. For example, some tasks naturally require architectural changes, while others make data-pipeline changes central. Thus, modification statistics should be interpreted as agent--task interactions rather than fixed agent properties.

\begin{figure}[t]
    \centering
    \includegraphics[width=0.98\linewidth]{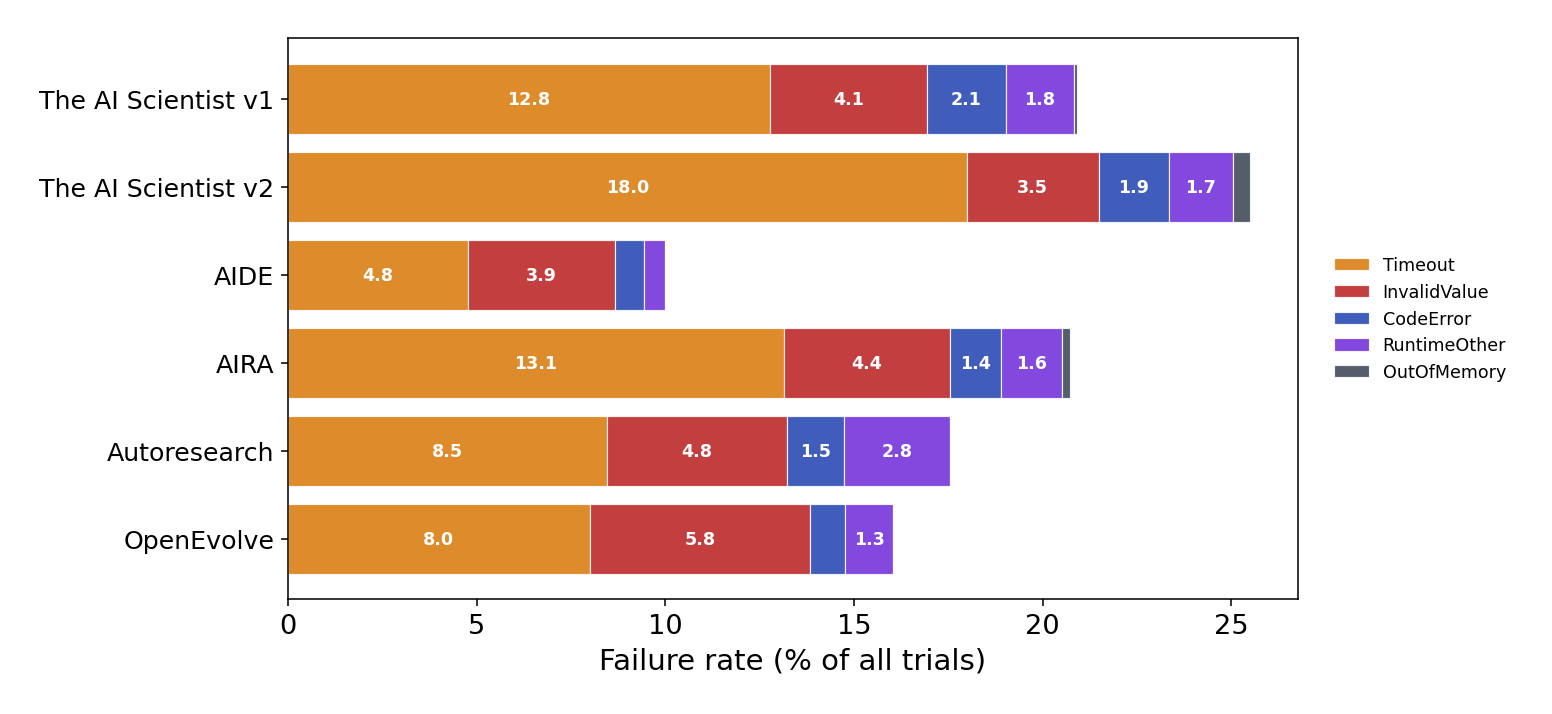}
    \caption{
    Failure-type rate for each agent, measured as the percentage of all trials ending in each failure type. Decomposing failures reveals whether an agent is timeout-bound, invalid-metric-prone, or affected by runtime implementation errors.
    }
    \label{fig:failure_type_rate_matrix}
\end{figure}

Figure~\ref{fig:failure_type_rate_matrix} decomposes failures by type. This view is more informative than a single aggregate failure rate because it separates distinct risk regimes. Some agents are timeout-dominated: their proposed experiments often run beyond the trial budget rather than failing immediately due to invalid code. Others are invalid-metric-prone: their code executes, but the final metric value is invalid after evaluation. A third regime consists of runtime or implementation errors, which can reveal mechanical bugs in how an agent edits code.

These failure regimes have different implications. Timeout-heavy agents may be proposing overly large or expensive interventions and would benefit from tighter edit control or a larger per-trial budget. Invalid-value-heavy agents need stronger internal checks on the expected metric output, since many of their failed trials complete execution but produce an invalid final metric. Runtime-error-heavy agents require targeted debugging of repeated implementation patterns. These distinctions would be hidden by a single failure-rate leaderboard.

Overall, agent strategy leaves a clear behavioral signature, but not through one scalar metric. The most useful diagnostic view combines what an agent chooses to modify with how those trials fail. Modification distributions reveal the agent's preferred search operator, while failure distributions reveal whether the resulting trials are too expensive, too weak, or mechanically invalid. Reporting both axes gives a more actionable picture of agent behavior than aggregate success or failure rates alone.

\section{Shared code editor}\label{app:codeeditor}

All six agents share a single code editor, implemented as a thin wrapper that issues one LLM API call per edit, logs the prompt and the produced patch, and applies the patch to the task repository under strict file-scope and safety constraints. The editor supports two output modes, selected automatically by file size: whole-file replacement for target files shorter than 500 lines, and line-anchored search-and-replace blocks for longer files. Agents pass natural-language instructions to the editor, but the editor embeds no agent-specific scaffolding, retrieval augmentation, or compiler feedback loop; there is no agent-specific intelligence in the editing layer.

Agents may modify only the files explicitly listed in each task's \texttt{target\_files} configuration. Every other file in the task repository, including evaluation scripts and dataset loaders, is structurally protected, which makes it architecturally impossible for an agent to tamper with evaluation code or ground-truth labels. A three-layer safety system enforces this contract in practice. First, every patch is screened by a static analyzer that blocks dangerous shell operations and arbitrary code execution. Second, each validation and test execution is run in an isolated environment in which the repository's version-control history is rendered read-only, preventing an agent's code from rewriting commit history or fabricating baseline snapshots. Third, a post-execution integrity check verifies that every protected file and the version-control history remain intact; if any layer fails, the step is recorded as a constraint violation and the change is rolled back. These mechanisms together guarantee that the code editor is an identical controlled tool for every agent, so that any performance gap reflects the agent's strategy rather than a difference in editing infrastructure.

\section{Representative task prompts}\label{app:prompts}

This appendix shows three representative task prompts, one per contrasting domain, to illustrate the prompt template used across the benchmark. Each prompt establishes the agent's role, the baseline algorithm, the optimization objective, and any task-specific constraints; the framework appends the current code, the metric display, and any prior-step history at run time.

\begin{tcolorbox}[colback=gray!10, colframe=black!50, boxrule=0.5pt, arc=2mm, title={Generalization (DomainBed, ColoredMNIST)}]
\small
\textbf{System:} You are an ambitious AI PhD student focused on improving generalization performance using the DomainBed benchmark.

\textbf{Task:} You are working with DomainBed's ERM method as the baseline on ColoredMNIST to evaluate generalization under distribution shifts.
Your goal is to enhance test-time domain generalization accuracy beyond standard ERM.
You should improve the algorithm based on ERM, but you may also propose entirely new algorithms if they better support cross-domain generalization.
The priority is to improve the average accuracy on unseen test domains while maintaining in-domain accuracy.
\end{tcolorbox}

\begin{tcolorbox}[colback=gray!10, colframe=black!50, boxrule=0.5pt, arc=2mm, title={Causality (gCastle, NOTEARS)}]
\small
\textbf{System:} You are an ambitious AI PhD student focused on causal structure learning and graph discovery.

\textbf{Task:} You are working with the NOTEARS algorithm for DAG structure learning on synthetic data.
Your goal is to reduce the structural Hamming distance (SHD) between the predicted and true causal graphs.
You may modify the NOTEARS optimization procedure, propose new structural constraints, or design entirely new DAG learning algorithms.
\end{tcolorbox}

\begin{tcolorbox}[colback=gray!10, colframe=black!50, boxrule=0.5pt, arc=2mm, title={Machine Unlearning (Open-Unlearning, TOFU)}]
\small
\textbf{System:} You are an ambitious AI PhD student focused on machine unlearning for large language models.

\textbf{Task:} You are working with the Gradient Ascent baseline on the TOFU benchmark built on Llama-3.2-1B. The forget-quality metric is the $p$-value of a Kolmogorov-Smirnov test between retained and forgotten outputs; to avoid numerical underflow at the baseline, we apply a post-hoc $-\log_{10}$ display transform, and your goal is to \emph{reduce} $-\log_{10}(p)$ (equivalently, raise the underlying $p$-value) while preserving model utility on the retain set. You may modify the unlearning loss, add regularization toward the retained distribution, or propose entirely new unlearning procedures.
\end{tcolorbox}

\section{Per-(agent, task) best-of-3 test improvement}\label{app:best_of_3}

Table~\ref{tab:normalized_performance} in the main text reports the \emph{mean-of-3} normalized test improvement for every (agent, task) cell. Table~\ref{tab:normalized_performance_best} below reports the \emph{best-of-3} counterpart on the same 18 tasks: for each cell, the value is the maximum over 3 rounds of the per-round normalized test improvement. This view answers a capacity / ceiling question (given 3 attempts, how good can each agent get on each task), and the gap to Table~\ref{tab:normalized_performance} quantifies the per-agent run-to-run upside.

\begin{table}[t]
\centering
\footnotesize
\caption{Normalized test improvement for each (agent, task) cell, taken as the \emph{best of 3} rounds. The final \textbf{MEAN over 18 tasks} row reports each agent's per-task best-of-3 mean and cross-task spread. Bold marks the per-row maximum (ties: all bolded).}
\label{tab:normalized_performance_best}
\begin{tabular}{lcccccc}
\toprule
Task & TAS v1 & TAS v2 & AIDE & AIRA & AutoR & OEvolve \\
\midrule
DomainBed-CM & 0.101 & 0.392 & 0.438 & 0.408 & 0.129 & \textbf{0.454} \\
DomainBed-OH & 0.009 & \textbf{0.015} & \textbf{0.015} & 0.013 & 0.008 & \textbf{0.015} \\
EasyFSL      & 0.036 & 0.053 & \textbf{0.084} & 0.077 & 0.063 & 0.077 \\
USB          & 0.047 & 0.048 & 0.038 & 0.024 & \textbf{0.081} & 0.079 \\
Lightly      & 0.049 & 0.036 & 0.070 & 0.000 & \textbf{0.101} & 0.057 \\
Solo-learn   & 0.096 & \textbf{0.117} & 0.092 & 0.101 & 0.006 & 0.099 \\
Cont.-Learn. & 0.060 & \textbf{0.727} & 0.299 & 0.215 & 0.641 & 0.231 \\
PyCIL        & 0.060 & \textbf{0.068} & 0.062 & 0.028 & 0.034 & 0.061 \\
CausalML     & 0.026 & \textbf{0.118} & 0.047 & 0.098 & 0.021 & 0.042 \\
gCastle      & 0.028 & 0.197 & \textbf{0.366} & 0.197 & 0.282 & 0.254 \\
ART          & 0.454 & 0.416 & 0.476 & 0.402 & \textbf{0.478} & 0.466 \\
OpenOOD      & 0.061 & 0.056 & 0.035 & 0.024 & \textbf{0.070} & 0.019 \\
PrivacyMeter & 0.024 & 0.540 & 0.550 & 0.034 & \textbf{0.597} & 0.097 \\
Opacus       & 0.000 & 0.000 & 0.018 & 0.000 & \textbf{0.070} & 0.005 \\
AIF360       & \textbf{0.239} & \textbf{0.239} & 0.234 & 0.226 & 0.233 & 0.238 \\
Fairlearn    & \textbf{0.173} & \textbf{0.173} & 0.164 & 0.155 & 0.171 & \textbf{0.173} \\
Unlearning   & 0.971 & 0.962 & \textbf{0.993} & 0.973 & 0.973 & 0.839 \\
PFLlib       & 0.248 & 0.251 & 0.257 & 0.273 & \textbf{0.338} & 0.207 \\
\midrule
\multirow{2}{*}{\textbf{MEAN over 18 tasks}}
& 0.149 & \textbf{0.245} & 0.235 & 0.180 & 0.239 & 0.190 \\
& $\pm$ 0.235 & \textbf{$\pm$ 0.268} & $\pm$ 0.256 & $\pm$ 0.237 & $\pm$ 0.270 & $\pm$ 0.212 \\
\bottomrule
\end{tabular}
\end{table}

\section{Per-agent process-metric Spearman correlations}\label{app:process_per_agent}

Table~\ref{tab:process_metrics} in the main text reports the pooled Spearman $\rho$ between each process metric and normalized test improvement over all 324 (round, agent, task) tuples. Table~\ref{tab:process_metrics_per_agent_rho} below reports the per-agent counterpart: for each (process metric, agent) cell, $\rho$ is computed over the 54 raw $(\text{round}, \text{task})$ tuples of that agent (3 rounds $\times$ 18 tasks pooled). This per-agent view exposes per-metric heterogeneity that the pooled row of Table~\ref{tab:process_metrics} averages out.

\begin{table}[t]
\centering
\footnotesize
\setlength{\tabcolsep}{2pt}
\caption{Per-agent Spearman $\rho$ between each process metric and normalized test improvement, computed over the 54 raw $(\text{round}, \text{task})$ tuples per agent (3 rounds $\times$ 18 tasks pooled). Significance: *** $p < 0.001$, ** $p < 0.01$, * $p < 0.05$. Anisotropy is omitted as redundant with Effective dim.}
\label{tab:process_metrics_per_agent_rho}
\begin{tabular}{llrrrrrr}
\toprule
Dimension & Metric & TAS v1 & TAS v2 & AIDE & AIRA & AutoR & OEvolve \\
\midrule
Exploration    & Exploration Spread       & $+$0.036    & $+$0.117    & $+$0.252      & $+$0.077     & $+$0.003    & $+$0.132 \\
Exploration    & Exploration Uniqueness   & $-$0.025    & $-$0.029    & $-$0.122      & $+$0.237     & $-$0.018    & $+$0.084 \\
Exploration    & Exploration Reach        & $+$0.014    & $+$0.072    & $+$0.329 *    & $+$0.011     & $+$0.078    & $+$0.193 \\
Exploration    & Effective dim            & $+$0.097    & $+$0.059    & $-$0.346 *    & $-$0.162     & $-$0.146    & $+$0.028 \\
Generalization & Val-test $|$gap$|$       & $-$0.000    & $+$0.228    & $+$0.118      & $+$0.358 **  & $-$0.011    & $-$0.135 \\
Reliability    & Valid step ratio         & $+$0.114    & $+$0.033    & $+$0.014      & $+$0.304 *   & $-$0.012    & $+$0.087 \\
Efficiency     & AUC-over-steps           & $+$0.694 *** & $+$0.913 *** & $+$0.830 *** & $+$0.711 *** & $+$0.866 *** & $+$0.700 *** \\
Efficiency     & First-improvement step   & $-$0.260    & $-$0.483 *** & $-$0.199     & $-$0.304 *   & $-$0.024    & $-$0.431 ** \\
Efficiency     & Late-gain fraction       & $-$0.020    & $-$0.238    & $+$0.162      & $-$0.182     & $-$0.012    & $-$0.123 \\
Efficiency     & Best-improvement step    & $+$0.055    & $-$0.117    & $+$0.117      & $-$0.088     & $+$0.182    & $+$0.177 \\
Cost           & Token cost (M)           & $+$0.036    & $+$0.123    & $+$0.072      & $-$0.190     & $-$0.013    & $-$0.018 \\
Cost           & Wall-clock time (h)      & $-$0.127    & $-$0.028    & $+$0.057      & $-$0.346 *   & $-$0.082    & $-$0.120 \\
\bottomrule
\end{tabular}
\end{table}

\section{Broader Impacts}\label{app:broader_impacts}

FML-bench is designed to evaluate and compare AI research agent strategies under controlled conditions. By providing process-level diagnostics alongside final performance, it can help the research community develop more effective and efficient agents, potentially accelerating progress across the ML domains covered by the benchmark (e.g., fairness, privacy, robustness). However, more capable AI research agents could also lower the barrier to conducting research with dual-use potential, such as generating adversarial attacks or circumventing privacy defenses. We note that all 18 tasks in FML-bench are built on publicly available research codebases and well-studied benchmarks. We encourage future users of FML-bench to consider the downstream implications of the research directions that automated agents may pursue.

\end{document}